\documentclass[conference]{IEEEtran}
\IEEEoverridecommandlockouts
\usepackage{cite}
\usepackage{amsmath,amssymb,amsfonts}
\usepackage{algorithmic}
\usepackage{graphicx}
\usepackage{subfigure}
\usepackage{textcomp}
\usepackage{tabularx}
\usepackage{makecell}
\usepackage{multirow}
\def\BibTeX{{\rm B\kern-.05em{\sc i\kern-.025em b}\kern-.08em
    T\kern-.1667em\lower.7ex\hbox{E}\kern-.125emX}}
\begin{document}

\title{\textbf{Object Detection for Autonomous Driving in Chinese Rural Scenes: An Experimental Study on Real-Synthetic Data Mixing and Model Evaluation}
}

\author{\IEEEauthorblockN{1\textsuperscript{st} Danning Zhu}
\IEEEauthorblockA{\textit{Wuhan University of Technology} \\
Wuhan, China \\
zhudanning.danny@whut.edu.cn}
\and
\IEEEauthorblockN{2\textsuperscript{nd} Ziyan Lin}
\IEEEauthorblockA{\textit{Wuhan University of Technology} \\
Wuhan, China \\
355843@whut.edu.cn}
\and
\IEEEauthorblockN{3\textsuperscript{rd} Jing Wu\textsuperscript{*}}
\IEEEauthorblockA{\textit{Wuhan University of Technology} \\
Wuhan, China \\
diamondwu@whut.edu.cn}
}

\maketitle

\begin{abstract}
Currently, autonomous driving object detection models face significant data scarcity and generalization challenges when navigating complex Chinese rural traffic scenarios. To address these limitations, we propose a novel real-synthetic mixed object detection dataset tailored specifically for Chinese rural roads, and systematically evaluate the performance of 13 mainstream detectors under different real-to-synthetic data ratios, thereby providing empirical evidence for model selection and data strategy design in rural autonomous driving scenarios. Our dataset combines real-world images captured in Weishi County, Henan Province, with parameterized virtual scenes generated via Unreal Engine. To accurately reflect the unique realities of rural traffic, we define a comprehensive 14-category object system encompassing region-specific elements such as electric tricycles, low-speed vehicles (LSVs), and roadside stalls. Under a unified training protocol, we systematically evaluate 13 mainstream detectors—spanning the YOLOv5, YOLOv8, YOLO11, and YOLO26 series, as well as RT-DETR-L—across three data configurations: an all-real baseline, a 1:0.5 real-to-virtual mix, and a 1:1 mix. Experimental results demonstrate that a moderate injection of synthetic data (1:0.5 ratio) effectively enhances detection performance, with YOLO11m achieving the highest mAP@0.5 of 0.758. However, a higher proportion of synthetic data (1:1) introduces domain shifts that offset the benefits of data scaling. While most models reliably identify distinct local vehicles, significant perceptual bottlenecks remain for long-tail, non-standard objects like stalls and railings. This research provides crucial empirical evidence and novel insights for model selection and synthetic data strategies, facilitating the practical deployment of autonomous driving perception systems in rural areas.

\end{abstract}


\section{Introduction}
Object detection is a core technical component of autonomous driving perception systems, and its performance directly determines the vehicle’s ability to perceive surrounding traffic participants and obstacles\cite{geiger2012we}. Single-stage detectors represented by the YOLO series, with their advantages of end-to-end inference and high real-time performance, have become a widespread choice for autonomous driving perception modules\cite{redmon2016you}\cite{redmon2017yolo9000}\cite{redmon2018yolov3}\cite{bochkovskiy2020yolov4}\cite{jocher2020yolov5}. Within this series, YOLOv5 introduces the CSPNet cross-stage partial network and PANet path aggregation mechanism, balancing accuracy and efficiency under a flexible model scaling strategy\cite{jocher2020yolov5}; YOLOv8 restructures the detection head design, adopting an anchor-free paradigm to simplify the decoding process and improve the recall of small objects; YOLO11 further optimizes the multi-scale feature representation of the backbone network and the label assignment strategy; and YOLO26, as the latest iteration, undergoes a comprehensive upgrade in network architecture, training data scale, and distillation strategy. In another direction, DETR\cite{carion2020end} introduces the self-attention mechanism of Transformers into object detection, achieving end-to-end set prediction via bipartite matching; RT-DETR\cite{zhao2024detrs} designs an efficient hybrid encoder to address the inference speed bottleneck, enabling Transformer-based detectors to achieve real-time inference speed comparable to the YOLO series for the first time. Hossain et al.\cite{hossain2025evaluating} conducted a systematic comparison of YOLOv8 and YOLO11 on a Bangladeshi urban traffic dataset, Alimov and Meiramkhanov \cite{alimov2024domain} evaluated the cross-domain generalization performance of YOLOv8s, RT-DETR, and YOLO-NAS in driving scenarios in Kazakhstan, and Schoder\cite{schoder2023first} performed a qualitative analysis of multiple YOLO generations and RT-DETR on diverse roads in Austria from a cautious driving perspective. These works provide valuable references for understanding the performance characteristics of models under road conditions in different regions, yet none of them has specifically conducted a systematic evaluation targeting Chinese rural scenes. The upper bound of object detection model performance depends on the coverage and diversity of training data\cite{sun2017revisiting}. Internationally mainstream autonomous driving datasets—KITTI \cite{geiger2013vision}, BDD100K \cite{yu2020bdd100k}, nuScenes \cite{caesar2020nuscenes}, and Waymo Open Dataset\cite{sun2020scalability}—are primarily collected in European and American urban environments; their object category systems fail to cover the unique composition of traffic elements on Chinese rural roads. The unique traffic participants and environmental characteristics of rural China cause models trained directly on these datasets to face severe out-of-domain generalization problems.

Chinese rural road scenes exhibit complexity distinct from urban environments in multiple dimensions. First, the composition of traffic participants is unique. As the primary means of short-distance passenger and freight transport in rural areas, electric tricycles possess non-standard dimensions and motion patterns, posing significant challenges for detectors trained on generic vehicle types to achieve accurate recognition. “\textit{Laotoule}” (low-speed electric four-wheel mobility vehicles, hereafter referred to as LSV) have experienced continuously growing ownership in villages and towns; their appearance characteristics, which fall between those of two-wheelers and cars, further increase classification ambiguity. In addition, temporary stalls at rural markets, goods piled along streets, and various forms of outdoor billboards constitute object categories that hardly exist in standard autonomous driving datasets, yet these objects have practical implications for safe path planning. Second, road infrastructure exhibits high heterogeneity. From paved county roads to unpaved field paths, there exists a continuous gradient variation in pavement material, width, alignment, and boundary clarity. Mixed pedestrian and vehicle traffic, along with missing or unclear traffic signs, is common in the core areas of villages and towns \cite{schoder2023first}\cite{cao2024semantic}\cite{yao2025construction}. These characteristics fundamentally differ from those of structured urban roads. Third, the geographic coverage required is extensive. China has a vast territory, and rural areas encompass diverse geomorphic and climatic types such as plain farming areas, hilly regions, and alpine pastoral areas, with significant regional differences in architectural styles, vegetation characteristics, and seasonal variations \cite{che2019d}. Real-world data captured from a single region are far from sufficient to train a detection model with cross-regional generalization capability.

Existing open-source datasets lack systematic coverage of Chinese rural scenes, and annotated data for objects such as tricycles, LSV, and stalls are particularly scarce. Within the broader scope of autonomous driving research, most benchmark datasets have focused on urban structured roads and highway scenarios \cite{wang2024m4sfwd}, leaving rural scenarios severely underrepresented. Among the limited related works, Yao et al. \cite{yao2025construction} constructed a rural road instance segmentation dataset in Xinjiang and evaluated the performance of multiple segmentation models on it. Cao et al. \cite{cao2024semantic} proposed an improved PP-LiteSeg network to address the semantic segmentation challenges of unstructured rural roads. In terms of datasets, although D²-City\cite{che2019d} and Rope3D\cite{ye2022rope3d} have introduced some categories with Chinese characteristics, the proportion of rural scenes and the number of categories remain insufficient to support the development of rural road perception systems. VisDrone\cite{zhu2021detection} collected data from a drone’s perspective across 14 cities in China, but it is oriented toward general traffic scenes and lacks a focus on the special objects present in rural areas. Accurate annotation of 14 object categories requires substantial professional labor, and expanding the geographic and seasonal coverage of data collection further drives up costs.

Synthetic data provides an important supplement for alleviating the issues of high annotation costs and insufficient long-tail scenario coverage in the autonomous driving domain\cite{voronin2024enhancing}. The generation methods of synthetic data can be broadly categorized into three classes. Graphics-based generation methods utilize game engines to construct virtual scenes and directly output pixel-level accurately annotated data: Damian et al. \cite{damian2023experimental} used Unreal Engine 5 in combination with domain randomization to generate training data, raising the F1 score of YOLOv8 from 0.15 to above 0.8; Voronin et al. \cite{voronin2024enhancing} employed the Unity engine to generate synthetic data for driving scenarios in New Zealand, where the real-plus-synthetic system surpassed the real-only system by approximately 3\% across all metrics; Kim et al. \cite{kim2024experimental} found that accuracy peaked at an 8:2 synthetic-to-real mixing ratio; and Lima et al. \cite{de2025optimizing} validated the effectiveness of synthetic pre-training combined with fine-tuning on a small amount of real data in maritime search and rescue scenarios. Generative adversarial network (GAN)-based methods generate augmented samples by learning the distribution of real data: Yao et al. \cite{yao2025construction} proposed an improved StyleGAN2-ADA for rural road scene data augmentation, raising the IS metric from 42.38 to 77.31 through a decoupled mapping network and convolutional coupling transfer blocks. Methods based on diffusion models and neural rendering represent the latest direction: Sim2Real Diffusion\cite{samak2025sim2real} utilizes conditional latent diffusion models to reduce the synthetic-to-real domain gap by over 40\%; Synth It Like KITTI\cite{marcus2025synth} verified the effectiveness of synthetic pre-training combined with fine-tuning on a small amount of real data. Regarding research on the mechanisms of synthetic data, Delussu et al.\cite{delussu2024synthetic} provided a methodological framework for the synthetic data research paradigm; Ljungqvist et al.\cite{ljungqvist2023object} found through CKA analysis that detectors trained on synthetic data exhibit highly consistent shallow-layer features with real data, but significant differences in the detection head layers, which holds reference value for understanding the mechanisms of knowledge transfer. Nevertheless, systematic experimental evidence across multiple generations of detection architectures and different real-to-synthetic data ratios is still lacking, leaving researchers without empirically supported references for model selection and data strategy design.

To address the above challenges, the main work of this paper are as follows. First, targeting the data scarcity problem of complex traffic scenes in rural China, we construct a hybrid real-synthetic object detection dataset containing 14 categories closely aligned with actual rural elements—including electric tricycles, LSV, and roadside stalls—by collecting real road images in Weishi County, Henan Province, and generating parametric virtual scenes using Unreal Engine. Second, under unified hyperparameter settings, we systematically evaluate the performance of 13 mainstream detectors—including YOLOv5, YOLOv8, YOLO11, YOLO26, and RT-DETR-L—under three data configurations: all-real, 1:0.5 real-to-synthetic mixed, and 1:1 real-to-synthetic mixed. Using experimental data, we directly compare the actual performance differences of multiple generations of detection architectures in rural road scenes. Third, the experiments deeply reveal the impact of different synthetic data ratios on the performance of various models, identify the optimal real-to-synthetic mixing ratio of 1:0.5, and highlight the performance degradation due to domain shifts caused by excessively high synthetic data ratios, the differences in sensitivity to real-to-synthetic domain shifts across architectures, and the perception bottleneck for long-tail non-standard objects, thereby providing crucial model selection and data strategy guidance for rural autonomous driving perception systems intended for real-world deployment.


\section{Dataset Construction }
This section provides an overview of the dataset construction process, covering real-world data acquisition, virtual scene generation, and dataset configuration and partitioning.

\subsection{Real-world Data Acquisition }

To build a foundational dataset that accurately reflects the actual traffic characteristics of rural China, rigorous real-world data collection was carried out. This subsection details the selection of the collection site, the definition of acquisition conditions, the categorization of scene types, and the design of the object category system and annotation protocol.

\textbf{Collection Location.}Real-world data were collected in several village and town areas within Weishi County, Kaifeng City, Henan Province, China. The selection of this site was based on considerations of representativeness across three dimensions. First, Weishi County is a typical agricultural county on the Central Plains of China. Its road network achieves full-spectrum coverage, ranging from paved county roads to unpaved field paths, offering a high degree of representativeness and strong potential for transferability to other rural areas in central China. Second, the area has high ownership rates of electric tricycles and LSVs, and roadside vendors and market activities are prevalent in the daily life of villages and towns, which accurately capture the core long-tail characteristics of Chinese rural traffic scenes. Third, the socio-economic indicators of Weishi County are at the median level among counties nationwide; data collected from this region represents, to a certain extent, the common infrastructure and traffic conditions found across China’s vast rural areas.
\begin{figure}[htbp]
    \centering
    \includegraphics[width=1\linewidth]{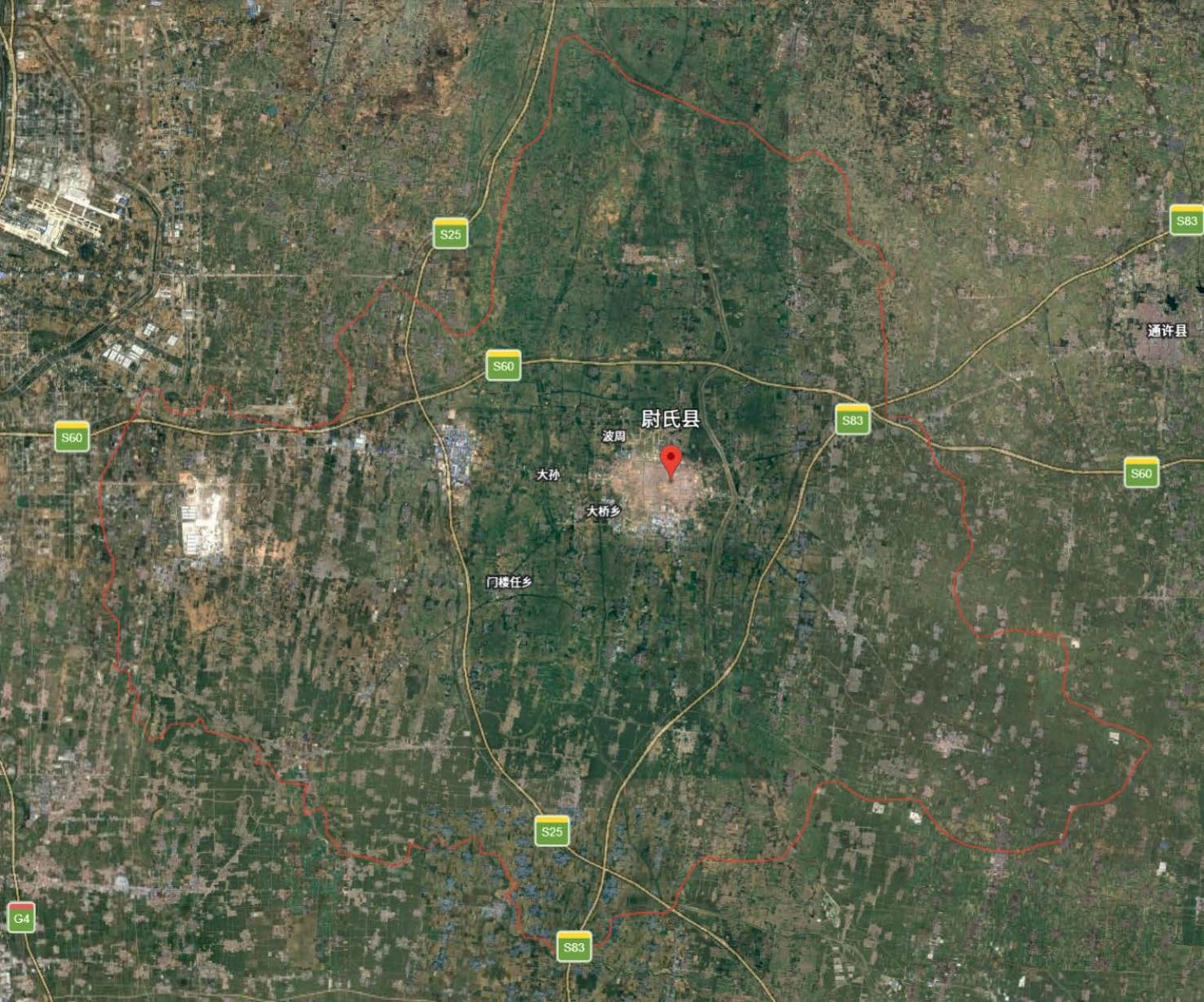}
    \caption{Satellite overview of Weishi County, Henan Province.}
    \label{fig:placeholder}
\end{figure}

\textbf{Acquisition Conditions.}Data collection was conducted in February 2026. To ensure the robustness of the model under varying illumination conditions, daily recording sessions were scheduled from 11:30 to 17:20, covering a range of natural lighting environments from high-illumination midday to low-illumination dusk. The acquisition device was a smartphone mounted at the center of the vehicle’s dashboard to simulate the true perspective of a vehicle-mounted front camera, ensuring a clear and unobstructed field of view. The video recording resolution was set to 1280×720 pixels at 30 FPS, with automatic exposure and automatic white balance enabled to approximate the actual optical parameters of a real vehicle camera. The cumulative effective recording time was approximately 144 minutes and 52 seconds, covering a driving distance of roughly 85 kilometers.

\textbf{Scene Types.}Considering the heterogeneity of rural road environments, the collected scenes were categorized into the following three types based on road class and environmental characteristics:

\textbf{1)County Roads:} Paved two-lane or two-way mixed-traffic roads. Lane markings exist but vary in clarity. The traffic flow includes a variety of types such as mini-trucks, agricultural tricycles, and electric scooters.

\textbf{2)Market Town Centers:} Core commercial road sections in market towns. These are characterized by high pedestrian density and a significant mix of motor vehicles, non-motor vehicles, and pedestrians. Roadside stalls and temporarily stacked goods occupy sidewalks and even parts of the roadway, resulting in low-structured traffic flow.

\textbf{3)Village Streets:} Internal roads within residential areas. These roads are narrow, flanked closely by residential courtyards, with pedestrians and livestock appearing randomly on the driving path. Vehicles typically need to pass through at low speeds and interact with residents.

\textbf{Object Category System.}To comprehensively cover the key elements in rural traffic scenes, we defined a system comprising 14 object categories. This system encompasses three major groups: dynamic traffic participants, static infrastructure, and key environmental background. The specific categories and their descriptions are shown in Table 1. Notably, scooters, tricycles, and LSVs are included with emphasis as high-frequency, non-standardized means of transport in rural areas. Stalls and billboards were also separately classified as rural-specific visual distractors and obstacles. This fine-grained category design aims to address the lack of rural element coverage in general autonomous driving datasets.

\begin{table*}
 \centering
 \footnotesize
\caption{Object categories.}
\begin{tabularx}{\textwidth}{p{4cm} p{4cm} l}
\hline
\textbf{CategoryID} & \textbf{Category Name} & \textbf{Remarks} \\
\hline
0 & scooter & Primary mode of individual transport tool in rural areas \\
1 & tricycle & Core tool for short-distance freight and passenger transport in rural areas \\
2 & LSV & Mobility vehicle typical of aging rural populations \\
3 & stall & Key obstacle affecting drivable area estimation \\
4 & bin & Common small roadside obstacle \\
5 & person & General object, often appearing in non-sidewalk areas in rural scenes \\
6 & car & General object, including sedans, SUVs, MPVs \\
7 & truck & Including various freight trucks and semi-trailers \\
8 & billboard & Visual distractor prone to trigger false detection/braking \\
9 & railing & Road boundary indicator \\
10 & pole & High-frequency infrastructure along rural roads \\
11 & street lamp & Present only on some road sections \\
12 & sign & General object, sparse and of diverse styles in rural scenes \\
13 & tree & Main environmental background element along roadsides \\
\hline

\end{tabularx}
\label{tab1}
\end{table*}

\textbf{Frame Extraction and Annotation.}To construct a high-quality image dataset from the video stream, a non-uniform temporal interval frame extraction strategy was applied. The extraction interval was adjusted according to the vehicle’s speed during recording—shorter intervals were used at higher speeds when the surrounding environment changes rapidly, and longer intervals were used otherwise—to reduce redundancy between adjacent frames and ensure sample diversity. A total of 4,720 static image samples were ultimately extracted. All objects appearing in each frame were annotated with rectangular bounding boxes using the xanylabeling tool, and the annotations were uniformly converted to the YOLO format after completion. A two-round cross-validation process was implemented to ensure annotation quality: the first round was completed independently by one annotator, and the second round involved a review and correction of all annotations by another annotator.
\begin{figure}[htbp!]
    \centering
    \subfigure[]{
    \begin{minipage}[b]{.4\linewidth}
        \centering
        \includegraphics[scale=0.1]{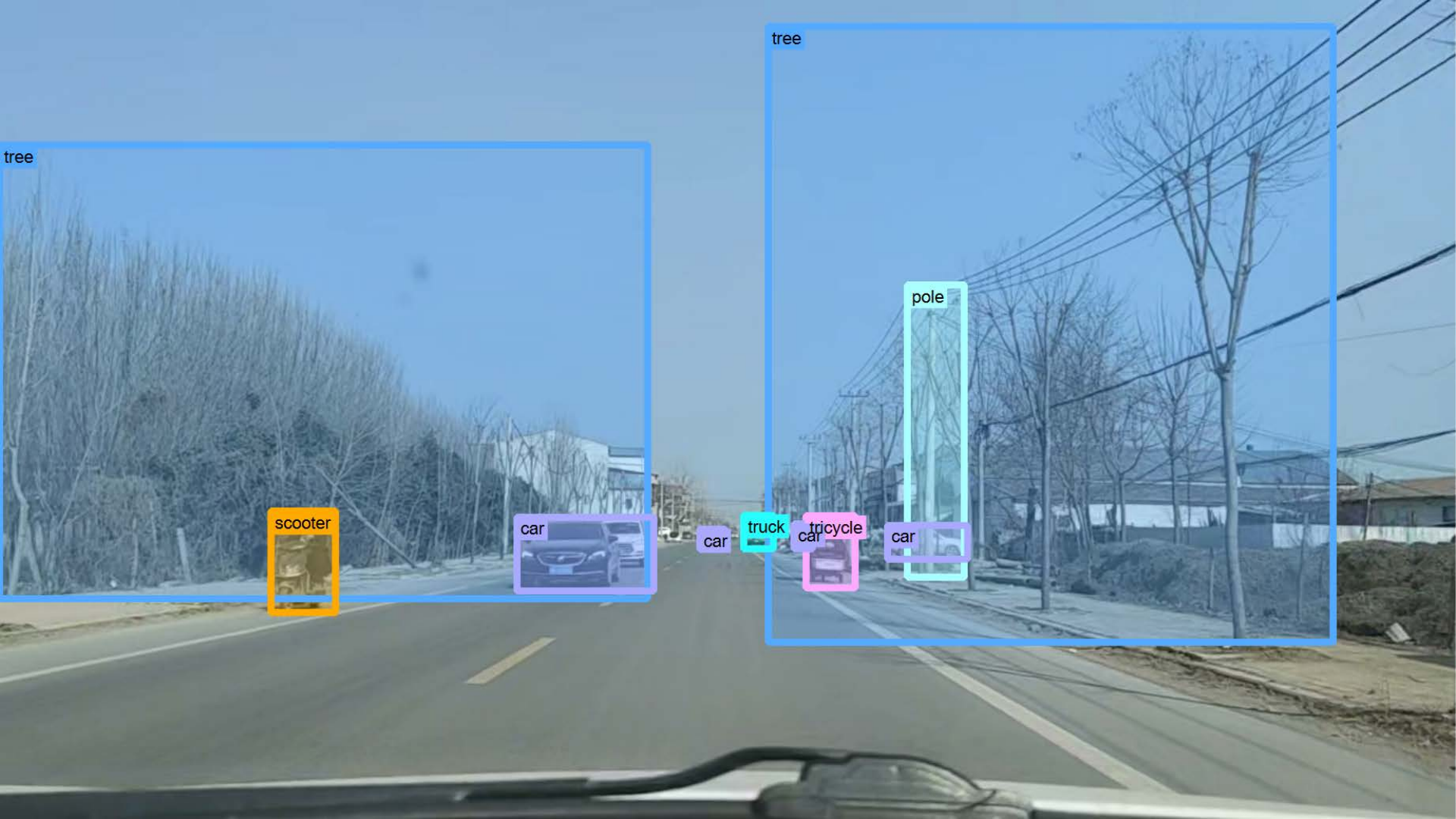}
    \end{minipage}
    }
    \subfigure[]{
    \begin{minipage}[b]{.4\linewidth}
        \centering
        \includegraphics[scale=0.1]{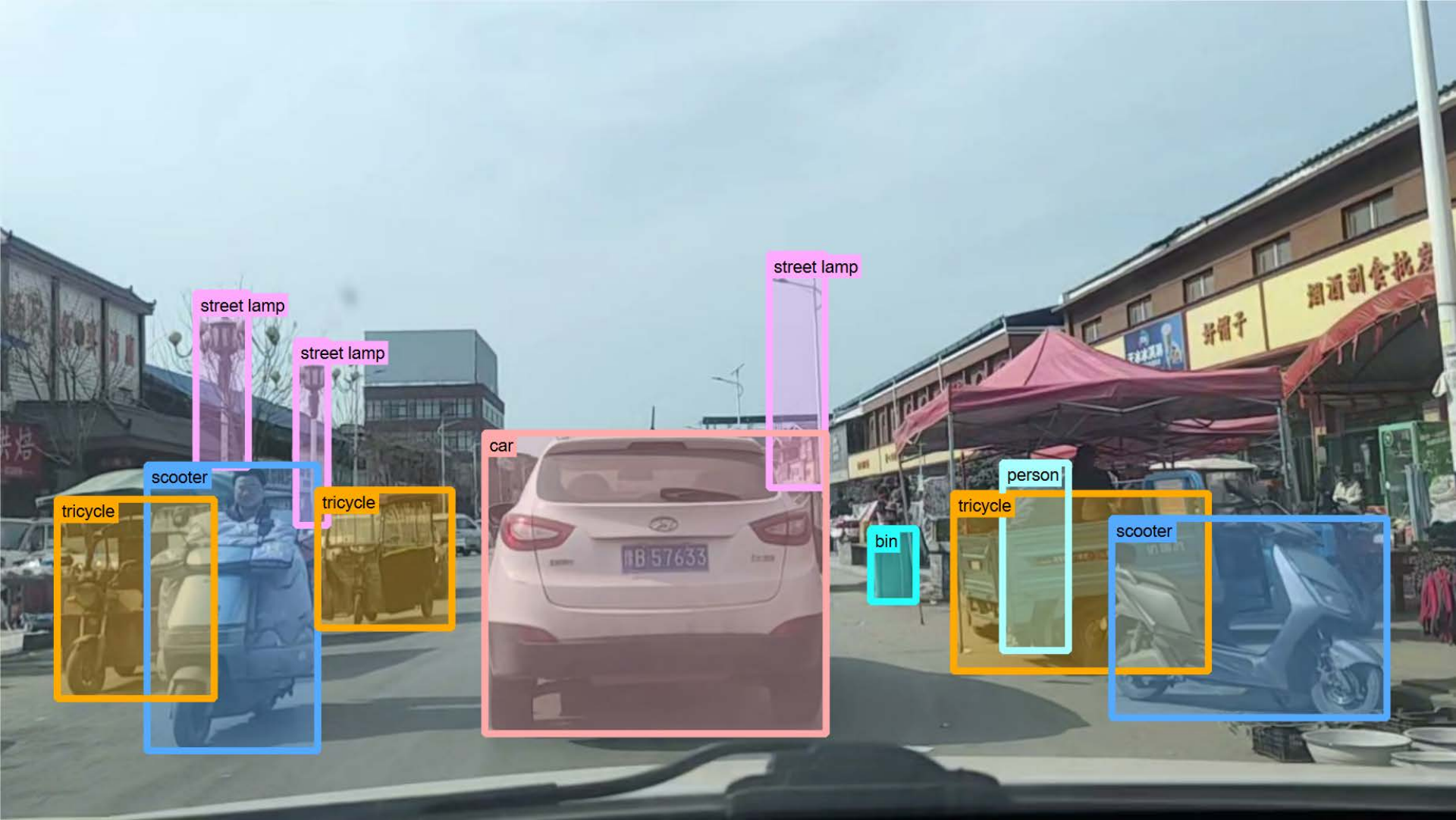}
    \end{minipage}
    }
    \subfigure[]{
    \begin{minipage}[b]{.4\linewidth}
        \centering
        \includegraphics[scale=0.1]{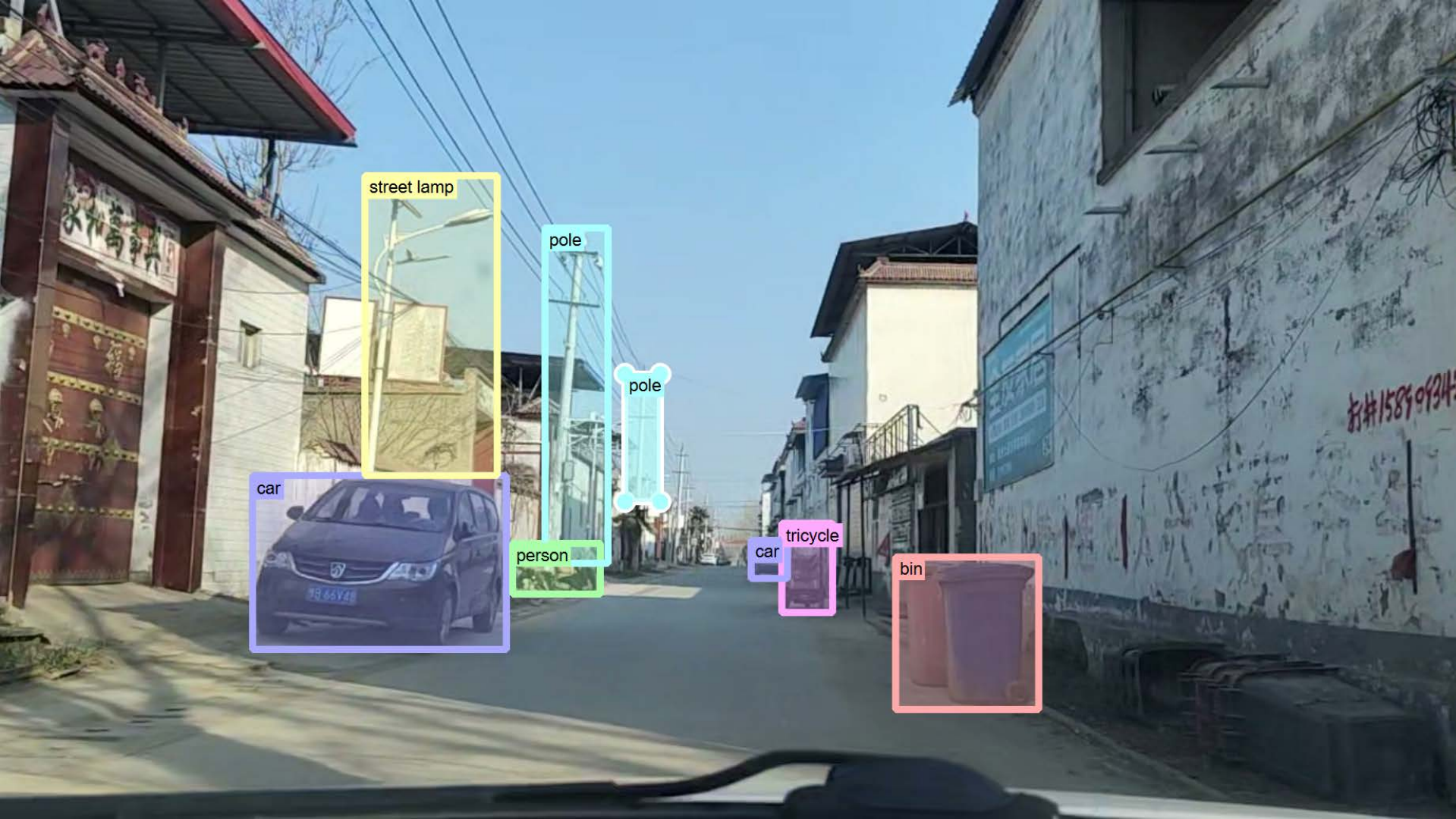}
    \end{minipage}
    }
    \caption{(a) County road, (b) Market town center, (c) Village street. }
    \label{fig:placeholder}
\end{figure}

\subsection{Virtual Scene Construction and Data Generation }
\textbf{Unreal Engine Platform.} In the selection of a synthetic data generation platform, mainstream options include Unity 3D and Unreal Engine. Compared to Unity, Unreal Engine offers comprehensive advantages in graphics rendering quality and the annotation tool ecosystem, making it the platform of choice for this study\cite{remmas2025pcgod}\cite{venkatesh2025volucapture}. The rationale for this choice can be further illustrated by three specific requirements of this research: for the need for high-fidelity image rendering, Unreal Engine 5.7’s Lumen global illumination and Nanite virtualized geometry technologies provide real-time, photorealistic image output; for the need for diverse scene construction, the Fab asset library and Blueprint scripting system support the rapid building and parametric adjustment of road scenes \cite{damian2023experimental}\cite{remmas2025pcgod}\cite{d2025syndra}; for the need for automated annotation, the EasySynth plugin, in conjunction with a self-developed conversion script, jointly realizes the conversion from semantic segmentation to YOLO labels \cite{venkatesh2025volucapture}\cite{d2025syndra}.

\textbf{Virtual Rural Scene Construction Process.}Chinese rural-style building models were imported from the Fab asset library and the Twinmotion marketplace. Infrastructure such as utility poles, streetlights, traffic sign poles, guardrails, and various outdoor billboards was simultaneously deployed, along with realistic road surface textures for rural paths, county roads, and village streets. Their density and spacing were set by referencing the effects observed in the real captured videos. High-precision 3D models of cars, scooters, tricycles, and LSVs were imported and placed with reference to real-world conditions to simulate realistic driving behavior along roads. The Citizen NPC system was used to generate pedestrian characters with diverse clothing and various actions, with irregular pedestrian trajectories simulated through random offsets of waypoints. Various types of roadside vendors, such as fruit stalls, snack stalls, and general merchandise stalls, were imported from Sketchfab and placed randomly along different types of roads by referencing real situations, simulating road occupancy states across different road sections.

\textbf{Data Acquisition and Label Generation.}A camera was set up in the Level Sequencer to move along a predefined path from a first-person perspective, with parameters kept consistent with the real-world acquisition: 1280×720 @ 30 FPS, f = 35.0 mm, aperture f/2.8, and a field of view of approximately 60°. The EasySynth plugin was then used to output semantic segmentation maps simultaneously with rendering the RGB images, such that objects of different categories were represented by distinct colors, and their contours were consistent with the actual object boundaries. To achieve the conversion from semantic segmentation to YOLO detection labels, a dedicated Python script, generate\_yolo\_labels.py, was written. This script parsed a CSV color mapping table to perform category-by-category color threshold binarization on the semantic segmentation map, applied morphological closing operations to repair breaks, extracted individual instances through connected-component analysis, calculated their minimum bounding rectangles, and finally output normalized detection labels in YOLO format.
\begin{figure}[htbp!]
    \centering
    \subfigure[]{
    \begin{minipage}[b]{.45\linewidth}
        \centering
        \includegraphics[scale=0.09]{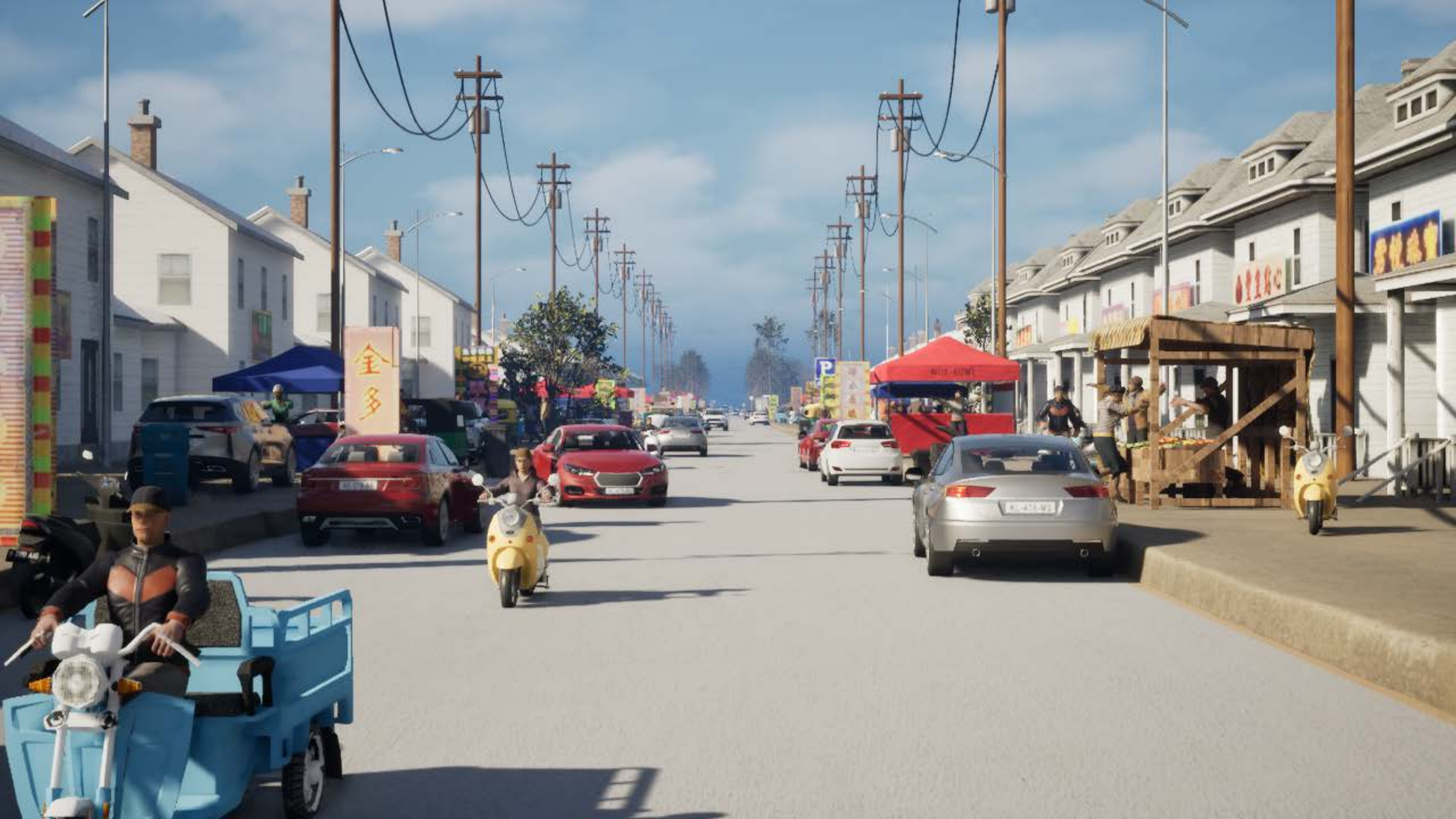}
    \end{minipage}
    }
    \subfigure[]{
    \begin{minipage}[b]{.45\linewidth}
        \centering
        \includegraphics[scale=0.09]{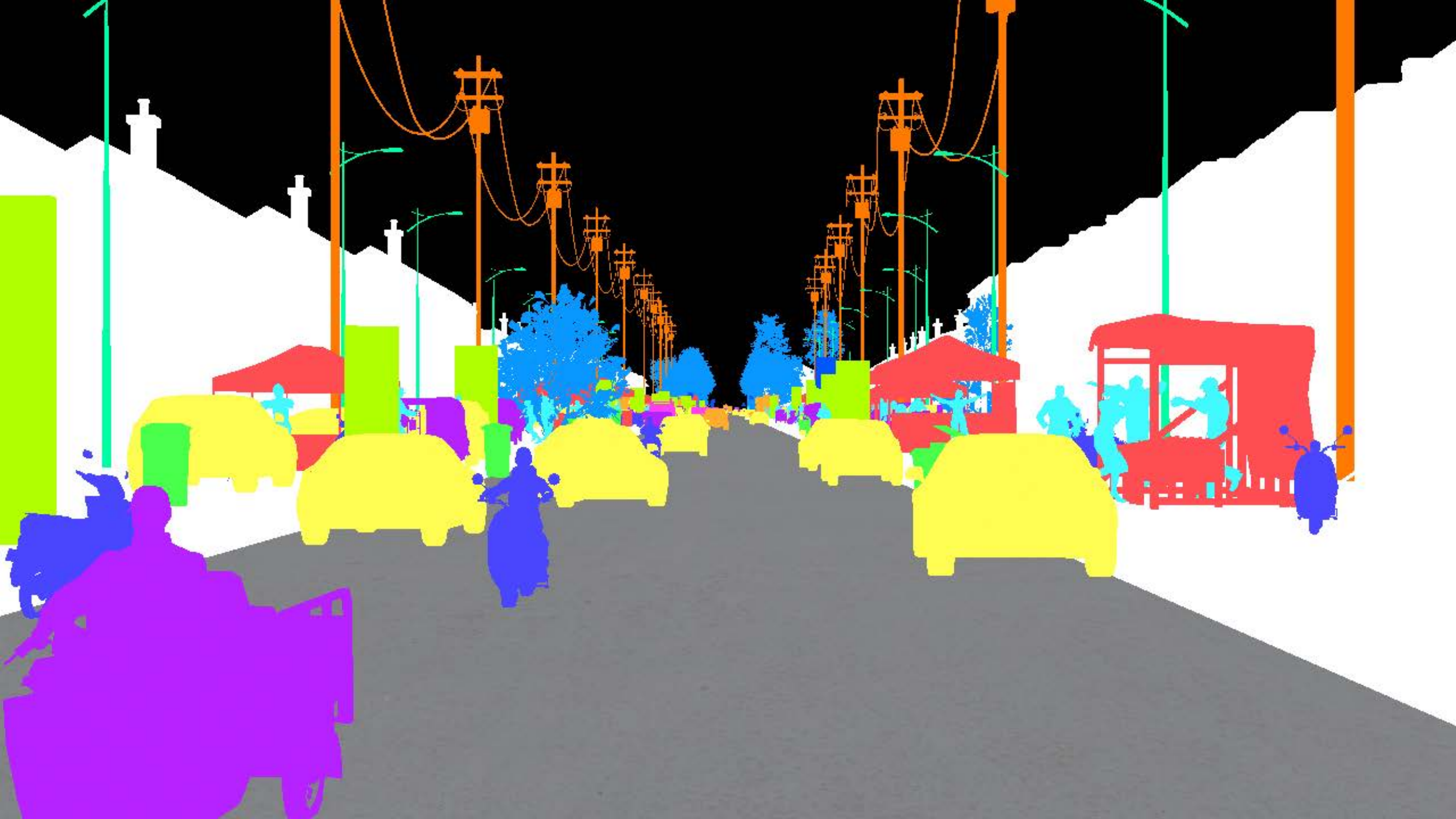}
    \end{minipage}
    }
    \subfigure[]{
    \begin{minipage}[b]{.45\linewidth}
        \centering
        \includegraphics[scale=0.09]{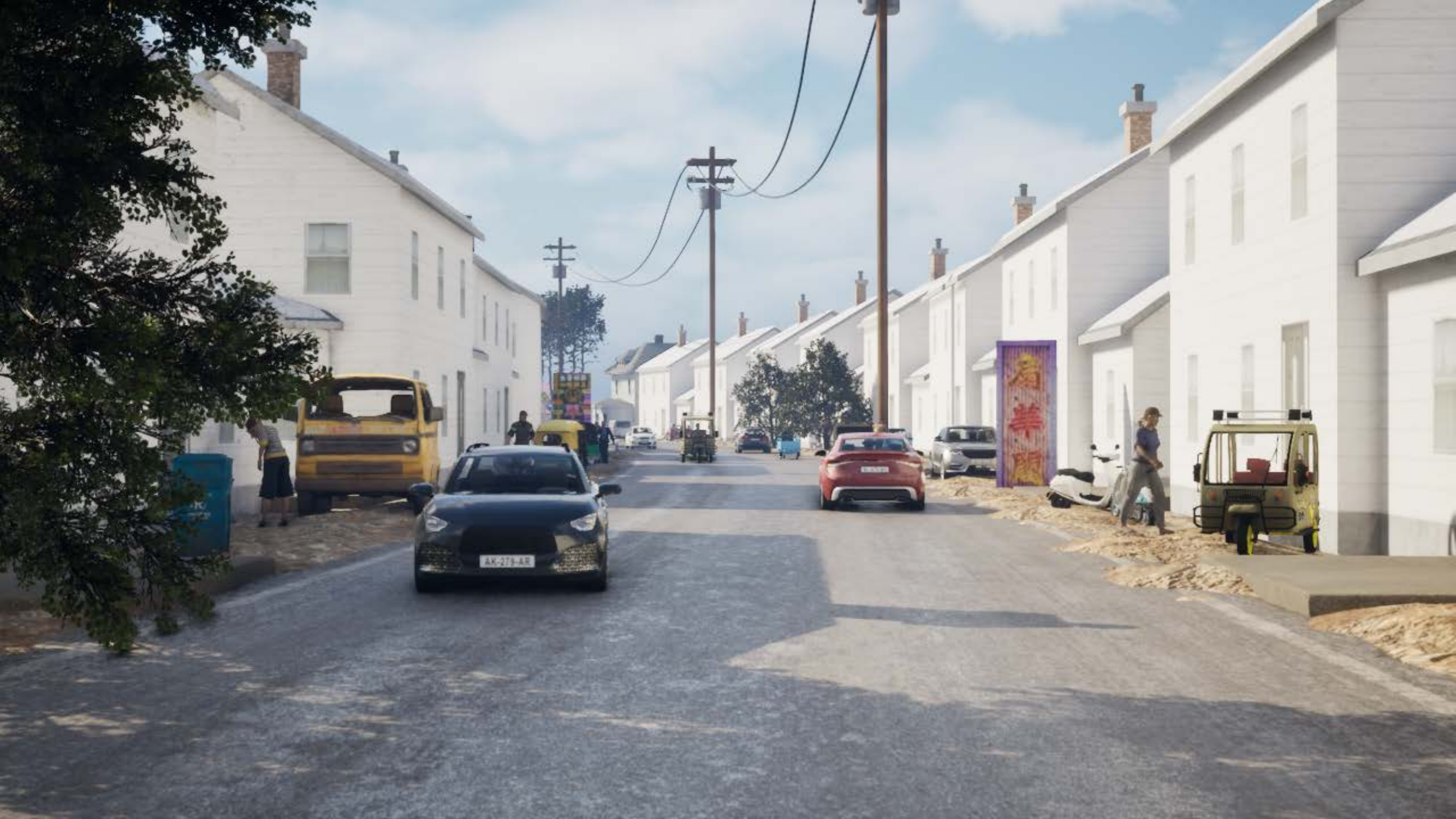}
    \end{minipage}
    }
    \subfigure[]{
    \begin{minipage}[b]{.45\linewidth}
        \centering
        \includegraphics[scale=0.09]{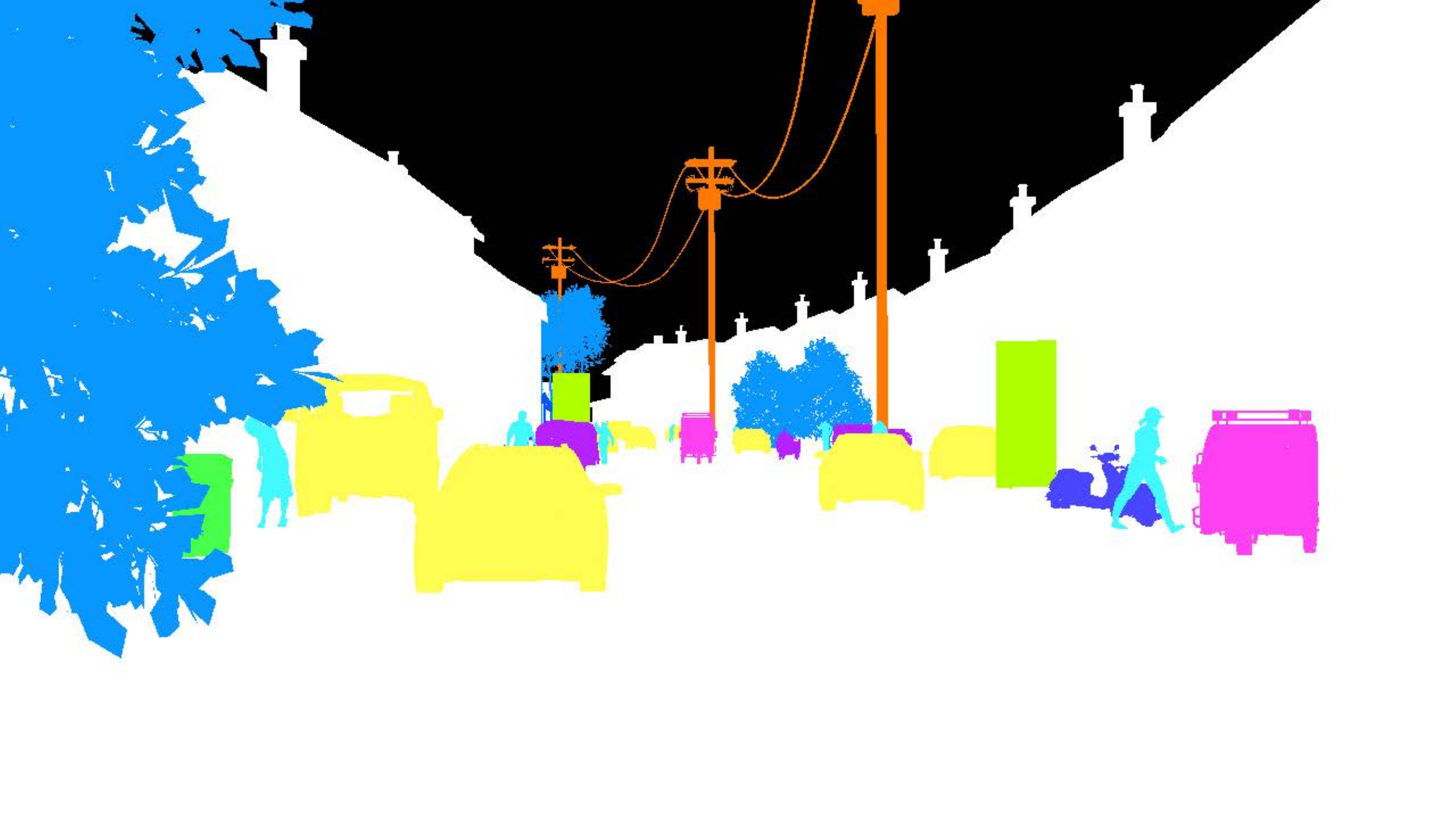}
    \end{minipage}
    }
    \subfigure[]{
    \begin{minipage}[b]{.45\linewidth}
        \centering
        \includegraphics[scale=0.09]{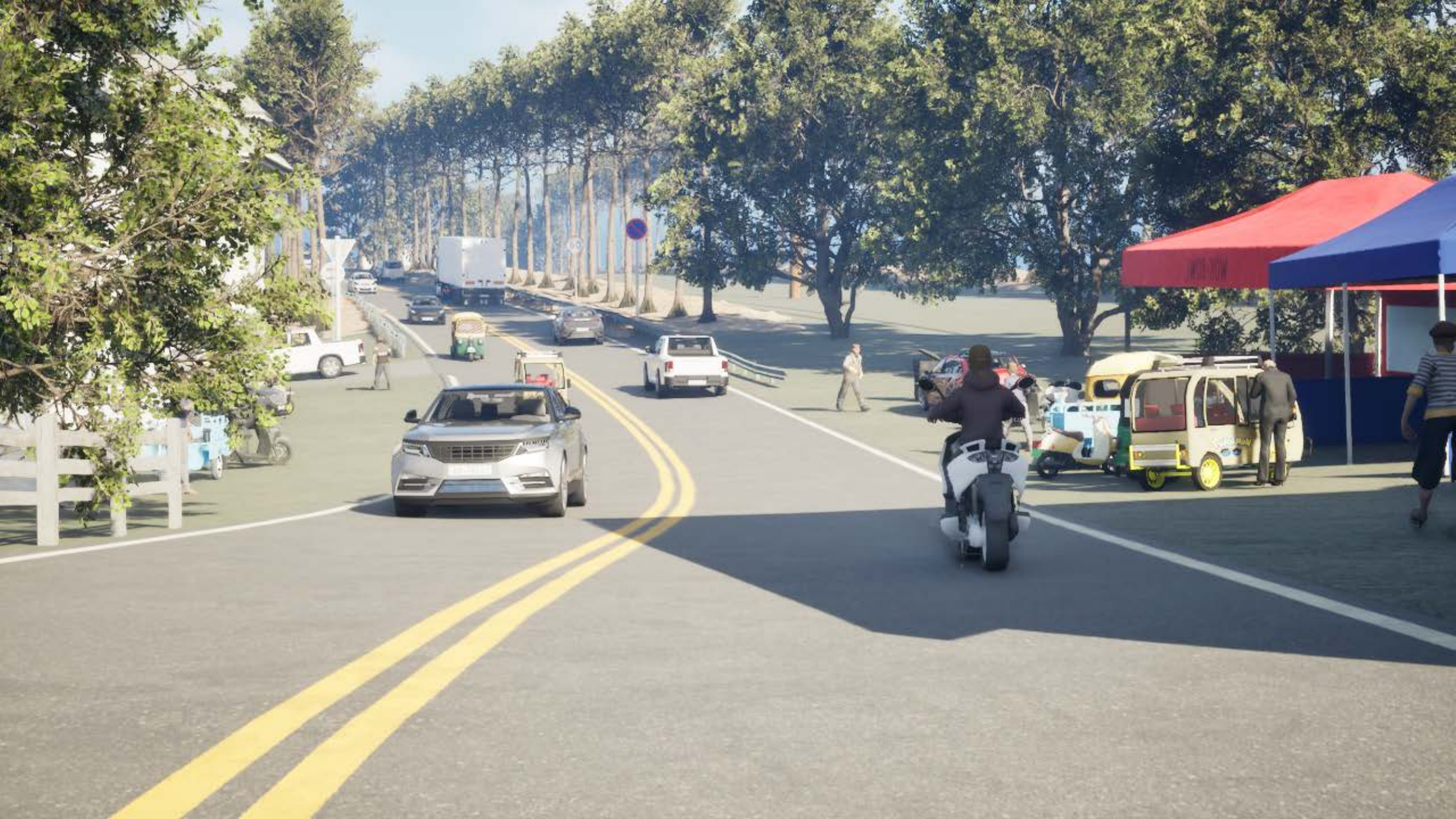}
    \end{minipage}
    }
    \subfigure[]{
    \begin{minipage}[b]{.45\linewidth}
        \centering
        \includegraphics[scale=0.09]{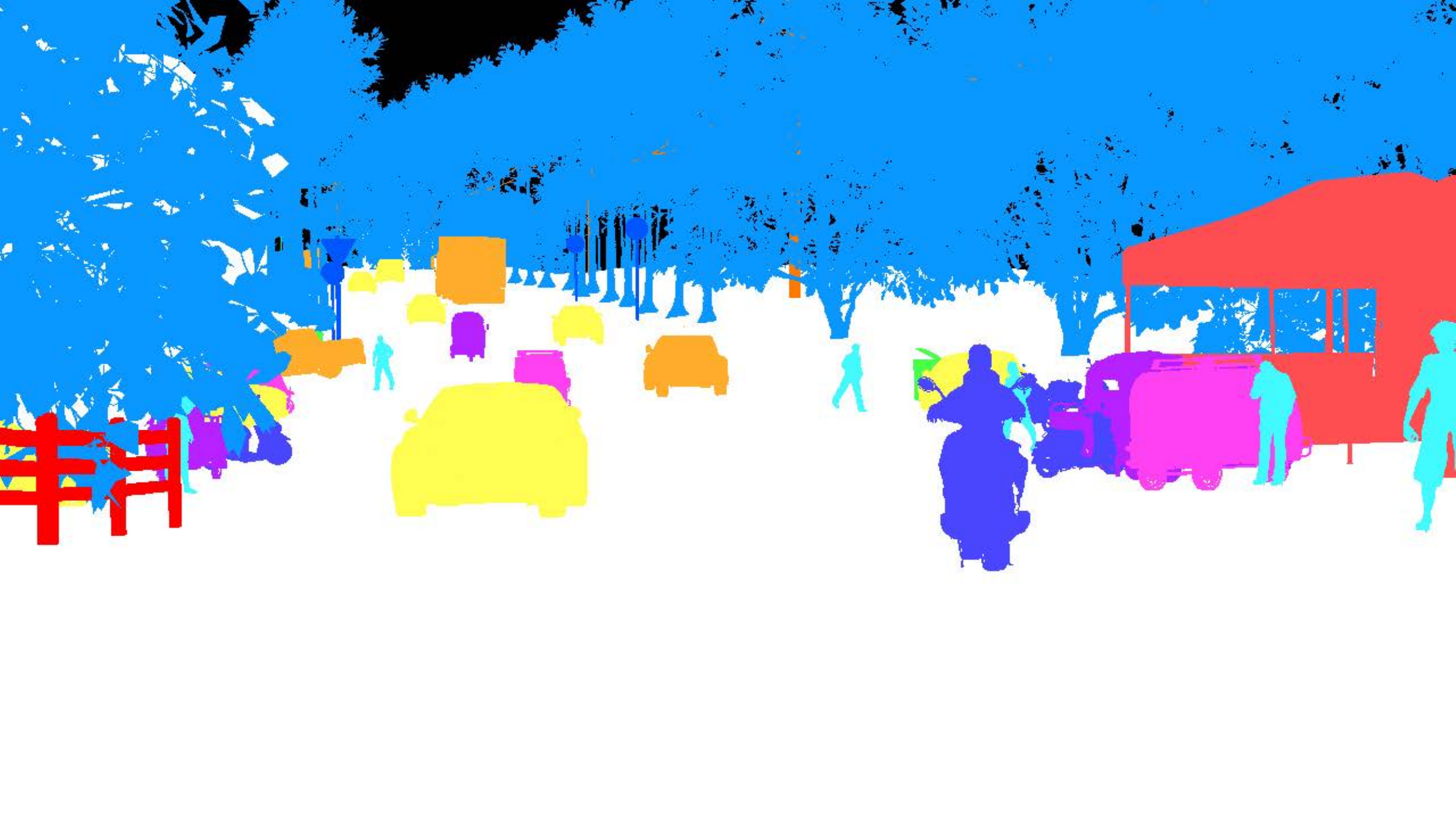}
    \end{minipage}
    }
    \caption{Virtual scenes of different road sections and their corresponding semantic segmentation maps.}
    \label{fig:placeholder}
\end{figure}

\subsection{Dataset Configuration and Partitioning }
To systematically investigate the impact of different real-to-synthetic data ratios on model performance, three configurations were constructed:

1) Configuration A——All-real: Only 4,200 real images were used.

2) Configuration B——Real : Synthetic = 1 : 0.5: 4,200 real images, supplemented by 2,100 synthetic images.

3) Configuration C——Real : Synthetic = 1 : 1: An equal number of real and synthetic images, with 4,200 images each.

All configurations were uniformly partitioned into training and validation sets at a 7:3 ratio. In particular, we set aside 517 real images as the test set. These images were carefully selected to include all 14 object categories, maximize environmental diversity as much as possible, and contained no synthetic data, ensuring that the evaluation results reflect the model’s actual generalization performance in real rural scenes.

\begin{table}
\centering
\caption{Sample distribution across the three dataset configurations.}
\begin{tabularx}{.48\textwidth}{llllc} 
\hline
         & real : synthetic &  Real Images & 
         \makecell[l]{Synthetic \\Images} & Total~  \\ 
\hline
Config A & 1:0               & 4200         & 0                 & 4200    \\
Config B & 1:0.5             & 4200         & 2100              & 6300    \\
Config C & 1:1               & 4200         & 4200              & 8400    \\
\hline
\end{tabularx}
\end{table}

\section{Experimental Design }
This section outlines the experimental setup designed to evaluate the proposed hybrid real-synthetic dataset for rural autonomous driving object detection. We first describe the selected models, evaluation metrics, experimental environment, and hyperparameter configuration.

\subsection{Model Selection }\label{AA}
A total of 13 detection models were selected for systematic comparison, covering two major architectural paradigms: CNN-based (YOLO series) and Transformer-based (RT-DETR). Specifically, four YOLO series—YOLO11, YOLO26, YOLOv8, and YOLOv5—were included, with three parameter scales (nano, small, and medium) chosen for each series. The parameter count of RT-DETR-L is 42M. The parameter sizes of the YOLO models are listed in Table 3.

\begin{table}[!t]
\centering
\setlength{\tabcolsep}{6pt}
\caption{Model parameters.}
\begin{tabularx}{.48\textwidth}{p{2cm} m{2cm} m{2cm} m{2cm}}
\hline
Model & \multicolumn{3}{c}{Parameters(M)} \\   
\hline
  & m & s & n \\
YOLO11 & 20.1 & 9.4 & 2.6 \\
YOLO26 & 20.4 & 9.5 & 2.4 \\
YOLOv8 & 25.9 & 11.2 & 3.2 \\
YOLOv5 & 25.1 & 9.1 & 2.6 \\
\hline
\end{tabularx}
\label{tab:model_params}
\end{table}

\subsection{Evaluation Metrics }
The following object detection metrics were adopted:

1) F1-score: Measures the harmonic mean between precision and recall, defined as F1 = 2 × (P × R) / (P + R), where precision P = TP / (TP + FP) indicates the accuracy of positive predictions, and recall R = TP / (TP + FN) reflects the completeness of true object detection.

2) mAP@0.5: The mean Average Precision computed over all classes at an IoU threshold of 0.5. This is the most commonly used comprehensive metric in practice.

\subsection{Experimental Environment }
Training was conducted on two computing platforms:

1) Platform 1 (Server): NVIDIA GeForce RTX 4090 (24 GB VRAM), driver version 580.105.08, CUDA 13.0. This platform was used for training tasks under Configuration B and Configuration C.

2) Platform 2 (Mobile Workstation): NVIDIA GeForce RTX 4060 Laptop (8 GB VRAM), driver version 581.57, CUDA 13.0. This platform was employed for training Configuration A and for inference testing of all models.

Differences in training efficiency between platforms (e.g., different batch sizes) did not affect the final model performance comparison, as all comparisons were based on the respective validation and test metrics rather than training speed.

\subsection{Hyperparameters }

To ensure that performance differences stem solely from model architecture and data configuration, all 13 models were trained under the three data configurations using unified default hyperparameters, without any model-specific or configuration-specific tuning. This established a fair controlled condition. The hyperparameter settings are detailed in Table 4.

\begin{table*}[t]  
    \centering
    \caption{Hyperparameter settings for model training.}
    \begin{tabularx}{\textwidth}{p{4cm} p{4cm} p{9cm}}  
        \hline
        \textbf{Hyperparameter} & \textbf{Value} & \textbf{Remarks} \\
        \hline
        Input image size & 640 × 640 & Letterbox resizing with aspect ratio padding \\
        Epochs & 100 & patience=100, equivalent to disabling early stopping \\
        Initial learning rate (lr0) & 0.01(YOLO) 1e-3(RT-DETR) & YOLO series uses SGD optimizer; RT-DETR uses AdamW optimizer, and its lr0 is automatically overridden to 1e-3 by the framework \\
        Final learning rate factor (lrf) & 0.01 & Linear decay; final learning rate is 1e-4 for YOLO series and 1e-5 for RT-DETR \\
        Momentum & 0.937 & SGD momentum \\
        Weight decay & $5\times10^{-4}$ & L2 regularization \\
        Box loss weight & 7.5 & CIoU loss for YOLO series; RT-DETR uses GIoU+L1 loss (this parameter does not apply) \\
        Cls loss weight & 0.5 & YOLO uses BCE; RT-DETR uses VariFocal Loss \\
        DFL loss weight & 1.5 & Distribution Focal Loss, YOLO series only; RT-DETR does not use DFL \\
        Mosaic augmentation & 1.0 (enabled) & Four-image mosaic; disabled in the last 10 epochs \\
        HSV augmentation & h=0.015, s=0.7, v=0.4 & Hue, saturation, value \\
        Horizontal flip probability & 0.5 & Random left-right flip \\
        Scale augmentation & 0.5 & Random scaling \\
        Translation augmentation & 0.1 & Random translation \\
        Pretrained weights & COCO official weights & All models are fine-tuned starting from COCO pretraining \\
        \hline
    \end{tabularx}
\end{table*}

\section{Experimental Results and Analysis }
Tables 5–7 and Fig. 4 summarize the two core metrics, mAP@0.5 and F1-score, of all models on the test set (517 real-only images) under the three data configurations.

\begin{table}
\centering
\caption{mAP@0.5 values of YOLO models on the test set.}

\begin{tabularx}{.48\textwidth}{p{0.5cm} p{0.4cm} p{0.4cm} p{0.5cm} p{0.4cm} p{0.4cm} p{0.5cm} p{0.4cm} p{0.4cm} p{0.5cm}}
\hline
model & \multicolumn{9}{c}{mAP@0.5} \\
\hline
\multirow{2}{*}{YOLO} & \multicolumn{3}{l}{Config A} & \multicolumn{3}{l}{Config B} & \multicolumn{3}{l}{Config C} \\
 & m & s & n & m & s & n & m & s & n \\
11 & 0.733 & 0.723 & 0.67 & 0.758 & 0.718 & 0.644 & 0.707 & 0.673 & 0.627 \\
26 & 0.707 & 0.694 & 0.621 & 0.708 & 0.665 & 0.621 & 0.71 & 0.666 & 0.61 \\
v8 & 0.714 & 0.712 & 0.668 & 0.736 & 0.72 & 0.667 & 0.699 & 0.677 & 0.624 \\
v5 & 0.747 & 0.714 & 0.656 & 0.743 & 0.704 & 0.661 & 0.704 & 0.671 & 0.633 \\
\hline

\end{tabularx}

\end{table}

\begin{table}
\centering
\caption{F1-scores of YOLO models on the test set.}
\begin{tabularx}{.48\textwidth}{p{0.5cm} p{0.4cm} p{0.4cm} p{0.5cm} p{0.4cm} p{0.4cm} p{0.5cm} p{0.4cm} p{0.4cm} p{0.5cm}}
\hline
model & \multicolumn{9}{c}{F1} \\
\hline
\multirow{2}{*}{YOLO} & \multicolumn{3}{l}{Config A} & \multicolumn{3}{l}{Config B} & \multicolumn{3}{l}{ConfigC} \\
 & m & s & n & m & s & n & m & s & n \\
11 & 0.67 & 0.65 & 0.62 & 0.7 & 0.65 & 0.6 & 0.65 & 0.62 & 0.59 \\
26 & 0.66 & 0.63 & 0.6 & 0.65 & 0.61 & 0.59 & 0.67 & 0.62 & 0.58 \\
v8 & 0.66 & 0.65 & 0.62 & 0.68 & 0.65 & 0.62 & 0.66 & 0.63 & 0.59 \\
v5 & 0.69 & 0.65 & 0.61 & 0.69 & 0.63 & 0.6 & 0.65 & 0.62 & 0.59 \\
\hline

\end{tabularx}

\end{table}

\begin{table}[t] 
\centering
\caption{mAP@0.5 and F1-score of RT-DETR-L on the test set.}
\begin{tabularx}{.48\textwidth}{p{2cm} p{1.8cm} p{1.8cm} p{2cm}}
\hline
 & Config A & Config B & Config C \\
\hline
mAP@0.5 & 0.736 & 0.692 & 0.691 \\
F1 & 0.68 & 0.66 & 0.66 \\
\hline

\end{tabularx}

\end{table}

\begin{figure}[htbp!]
    \centering
    \subfigure[]{
    \begin{minipage}[b]{.4\linewidth}
        \centering
        \includegraphics[scale=0.15]{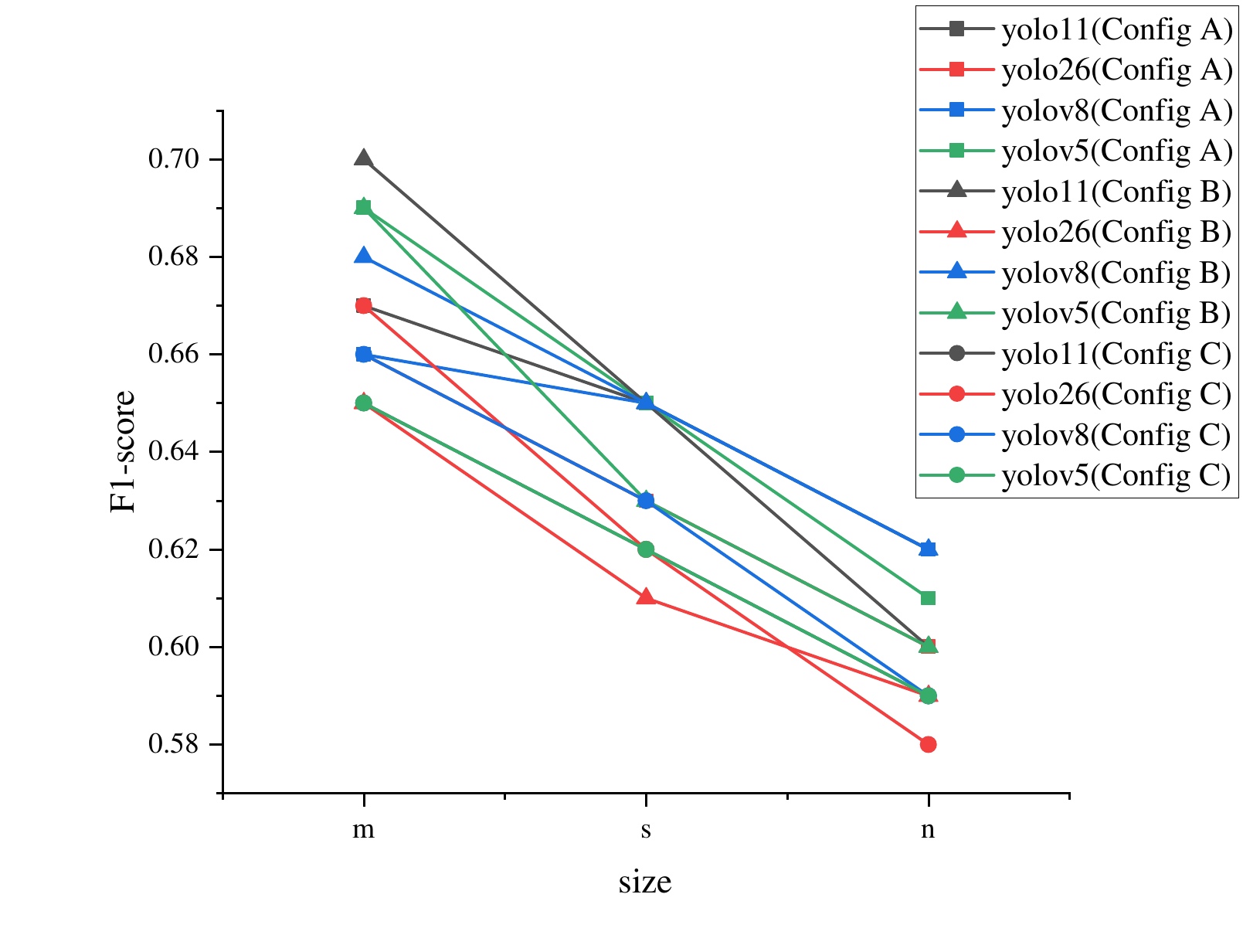}
    \end{minipage}
    }
    \subfigure[]{
    \begin{minipage}[b]{.4\linewidth}
        \centering
        \includegraphics[scale=0.15]{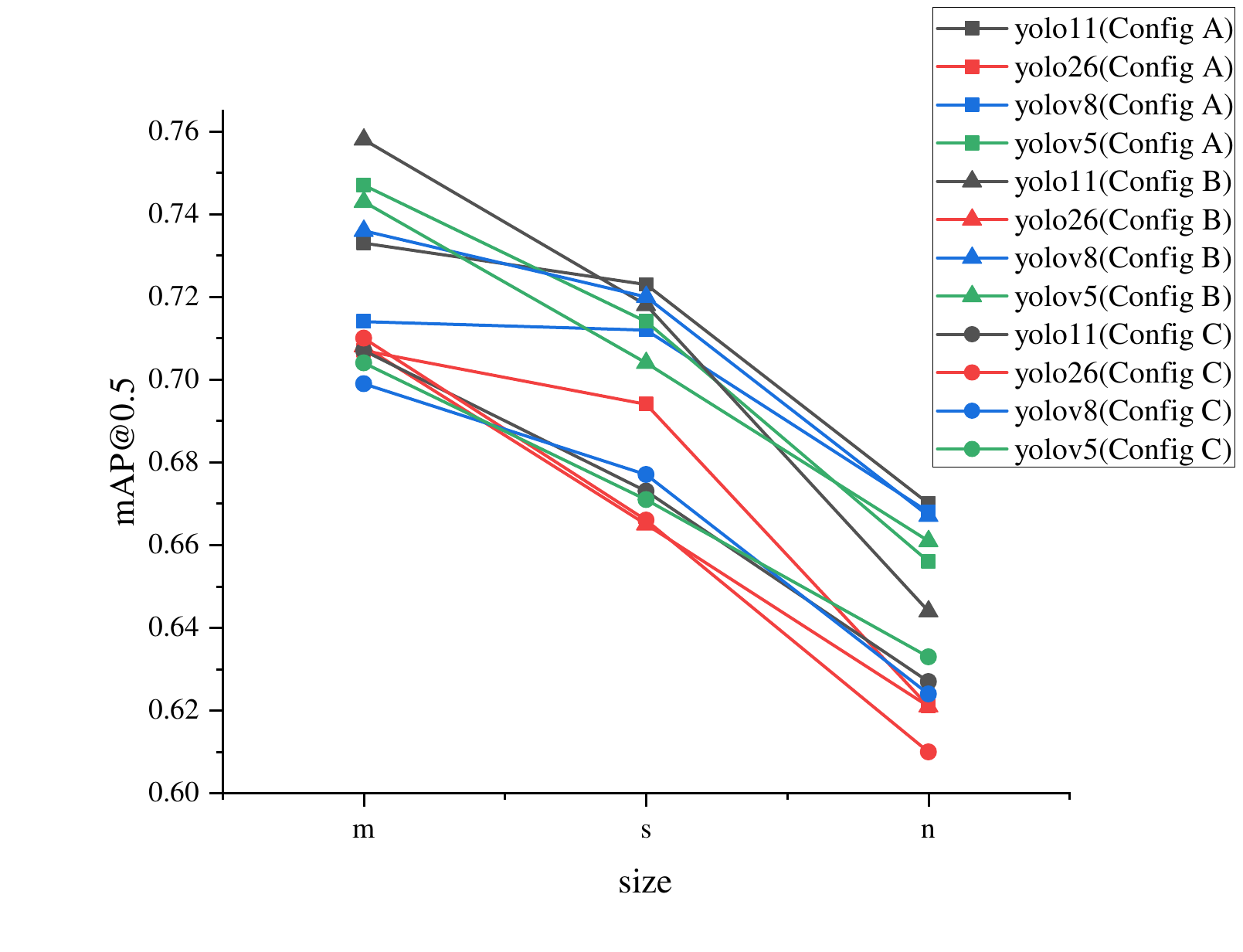}
    \end{minipage}
    }
    \caption{mAP@0.5 and F1-score of different YOLO model sizes and configurations on the test set. }
    \label{fig:placeholder}
\end{figure}

The experimental results demonstrate that a certain amount of synthetic data contributes positively to model performance. As can be seen from the line plots in Fig. 4, for both mAP@0.5 and F1-score, the curves under Configuration B (indicated by triangles) mostly lie at higher levels compared to other configurations. The highest mAP@0.5 and F1-score are both achieved by YOLO11m under Configuration B, with an mAP@0.5 of 0.758, representing an improvement of 0.025 over Configuration A, and an F1-score of 0.7, an improvement of 0.03 over Configuration A. YOLOv5mu under Configuration A exhibits the next best performance. In contrast, models under Configuration C (indicated by circles) perform notably worse than those under the other two configurations, with the poorest performance observed for YOLO26n under Configuration C, achieving an mAP@0.5 of 0.61 and an F1-score of 0.58.

From Fig. 4, a clear positive correlation between model parameter scale and detection accuracy is observed under the unified hyperparameter setting. Comparing models of n, s, and m sizes across the YOLO series reveals that increasing the scale consistently leads to performance gains. Taking YOLO11 under Configuration B as an example, the mAP@0.5 values for the n, s, and m scales are 0.644, 0.718, and 0.758, respectively, with the medium-scale model achieving an 11.4 percentage point improvement over the nano-scale model. This trend is similarly observed in YOLOv8, YOLO26, and YOLOv5. This suggests that Chinese rural traffic scenes contain numerous objects with highly variable appearances, such as modified vehicles and temporary stalls, which require backbone networks with higher parameter capacity for complex spatial feature modeling; models that are too small tend to encounter representation bottlenecks in such complex scenes.

Fig. 5 and Table 8 present the test set mAP@0.5 of the medium-scale YOLO models and RT-DETR-L under the three configurations. Analysis of these data reveals distinct performance patterns for different model series under varying synthetic data ratios.

YOLO11m and YOLOv8m both exhibit an inverted U-shaped trend: YOLO11m rises from 0.733 under Configuration A to 0.758 under Configuration B, then drops to 0.707 under Configuration C; YOLOv8m increases from 0.714 under Configuration A to 0.736 under Configuration B, then drops to 0.699 under Configuration C. This pattern indicates that a moderate amount of synthetic data plays a positive role in regularization and scene diversification for YOLO11m and YOLOv8m, whereas excessive synthetic data introduces too much domain bias, ultimately harming model performance on the real test set. This phenomenon aligns with the observation of Lima et al. \cite{de2025optimizing} in maritime search and rescue scenarios, namely, that there exists an optimal real-to-synthetic mixing ratio, beyond which the gains diminish or even turn negative.

For YOLOv5mu and RT-DETR-L, however, the introduction of synthetic data leads to performance degradation. YOLOv5mu drops from 0.747 under Configuration A to 0.743 under Configuration B, and then further to 0.704 under Configuration C. RT-DETR-L exhibits a substantial decline from 0.736 under Configuration A to 0.692 under Configuration B, followed by a slight decrease to 0.691 under Configuration C. These results indicate that the architecture of YOLOv5mu is relatively sensitive to the domain shift introduced by synthetic data, with a small amount of synthetic data providing no positive effect and performance declining noticeably as more synthetic data are added. RT-DETR-L, which adopts a Transformer architecture differing from the CNN structure of YOLO models, experiences a marked performance drop even with a small amount of synthetic data. This may be attributed to the global attention mechanism of RT-DETR, which makes it more dependent on the overall statistical properties of the training data distribution; the discrepancies in texture, illumination, and shadow distribution between synthetic and real data are amplified in the computation of attention weights.

Notably, YOLO26m shows almost no performance variation across the three configurations. This may be because YOLO26 employs an end-to-end one-to-one detection head design, eliminating the dependence on NMS post-processing, so its predictions are not affected by fluctuations in data distribution. Additionally, it removes the DFL loss and introduces a progressive loss strategy, simplifying the regression objective and reinforcing direct optimization of output quality, thereby significantly reducing sensitivity to differences in training data composition.
\begin{figure}[htbp]
    \centering
    \includegraphics[width=1\linewidth]{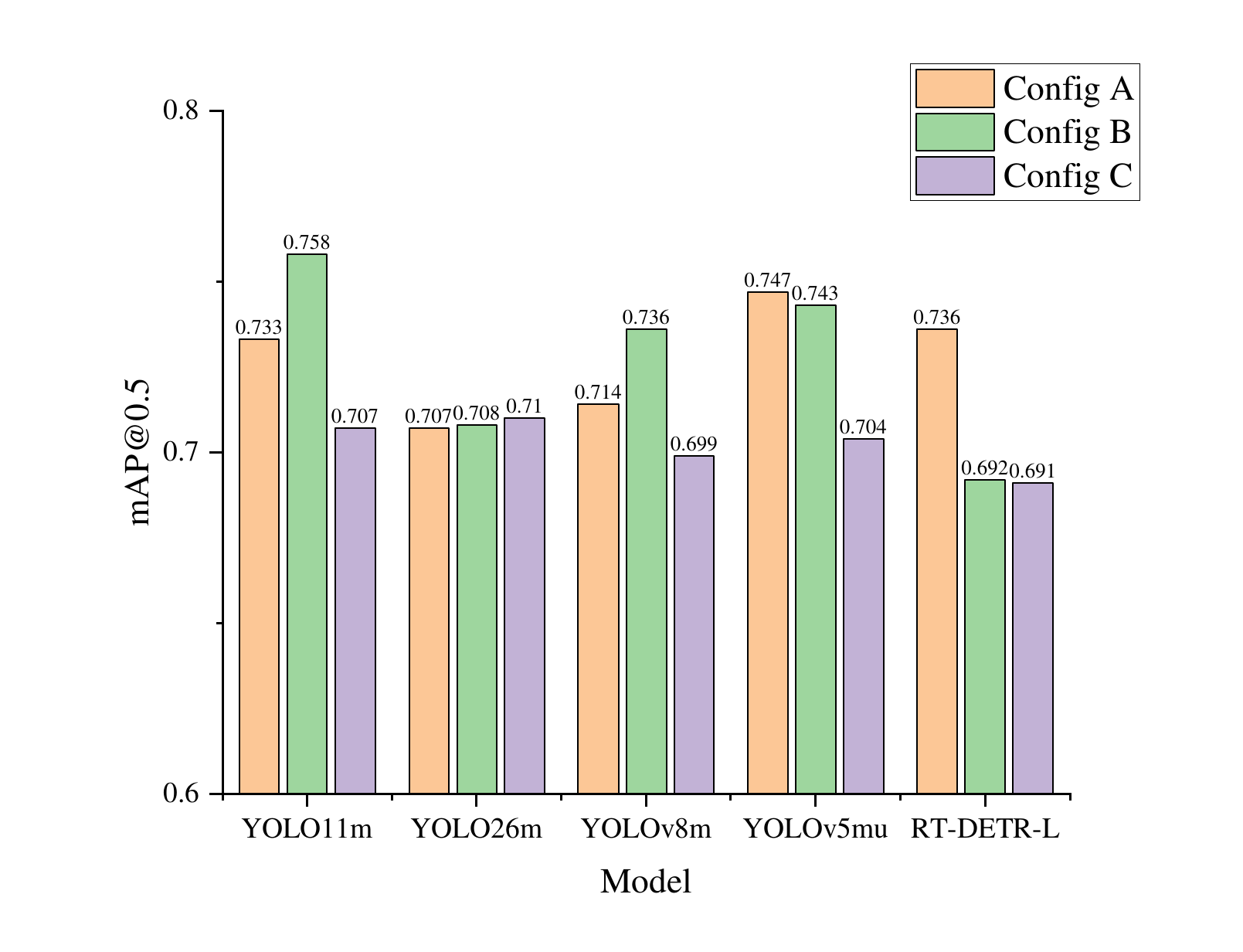}
    \caption{mAP@0.5 of medium-scale YOLO models and RT-DETR-L under the three configurations.}
    \label{fig:placeholder}
\end{figure}

\begin{table}
\centering
\caption{mAP@0.5 of medium-scale YOLO models and RT-DETR-L under the three configurations.}
\begin{tabularx}{.48\textwidth}{m{0.6cm} m{1.1cm} m{1.1cm}m{1.1cm} m{1.4cm}m{1cm}}
\hline
Config & YOLO11m & YOLO26m & YOLOv8m & YOLOv5mu & RT-DETR-L \\
\hline
A & 0.733 & 0.707  & 0.714 & 0.747 & 0.736 \\
B & 0.758 & 0.708 & 0.736 & 0.743 & 0.692 \\
C & 0.707 & 0.71 & 0.699 & 0.704 & 0.691 \\
\hline

\end{tabularx}

\end{table}

Table 9 and Fig. 6 present the per-class mAP@0.5 of all models under Configuration B, which delivers the best overall performance. A closer analysis reveals a pronounced polarization in the feature extraction and recognition capabilities of different models across the 14 categories.

First, mainstream models all achieve extremely high detection reliability on common objects such as cars (mAP@0.5 approximately 0.92–0.95) and trash bins (approximately 0.83–0.87). More importantly, for the highly representative categories of Chinese rural scenes—tricycles and LSVs—the models also achieve excellent results. YOLO11m attains a tricycle mAP@0.5 of 0.879, and YOLOv5mu reaches 0.877 on LSVs (significantly outperforming RT-DETR-L’s 0.797). This strongly validates that the proposed hybrid real-synthetic dataset can provide sufficient and high-quality training samples for these characteristic traffic participants, enabling the models to fully learn their typical appearance features and substantially compensating for the generalization deficiency of general autonomous driving datasets in rural scenarios.

However, stalls, trucks and railings remain significant long-tail challenges in current perception tasks. The performance of all models on stalls is generally limited; although YOLOv5mu (0.589) and YOLO11m (0.560) exhibit relatively better performance, the overall mAP@0.5 mainly hovers between 0.45 and 0.59. This is primarily because rural market stalls are extremely non-standard in appearance, highly variable in shape, and often accompanied by severe self-occlusion and mutual occlusion caused by dense crowds and goods.

The detection results for the railing category are the weakest. This is likely because railings appear infrequently in the test set and exhibit highly variable forms; they are typically underrepresented in real-world datasets. Moreover, railings are typical elongated objects, and in the background they often overlap with distractors with highly similar textures such as trees and utility poles, easily causing visual feature confusion. In this category, the best-performing YOLO11m achieves an mAP@0.5 of only 0.470, YOLOv5mu drops to 0.211, and the Transformer-based RT-DETR-L even suffers severe detection failure with an mAP@0.5 of merely 0.024. This indicates that current mainstream architectures, whether based on the local convolution kernels of conventional CNNs or the global self-attention mechanism (e.g., RT-DETR), still exhibit clear representation bottlenecks when performing bounding box regression for such elongated objects with variable appearances, data scarcity, and complex backgrounds. It underscores the urgent need to improve the perception capability for non-standard obstacles in unstructured rural scenes.
\begin{figure}[htbp]
    \centering
    \includegraphics[width=1\linewidth]{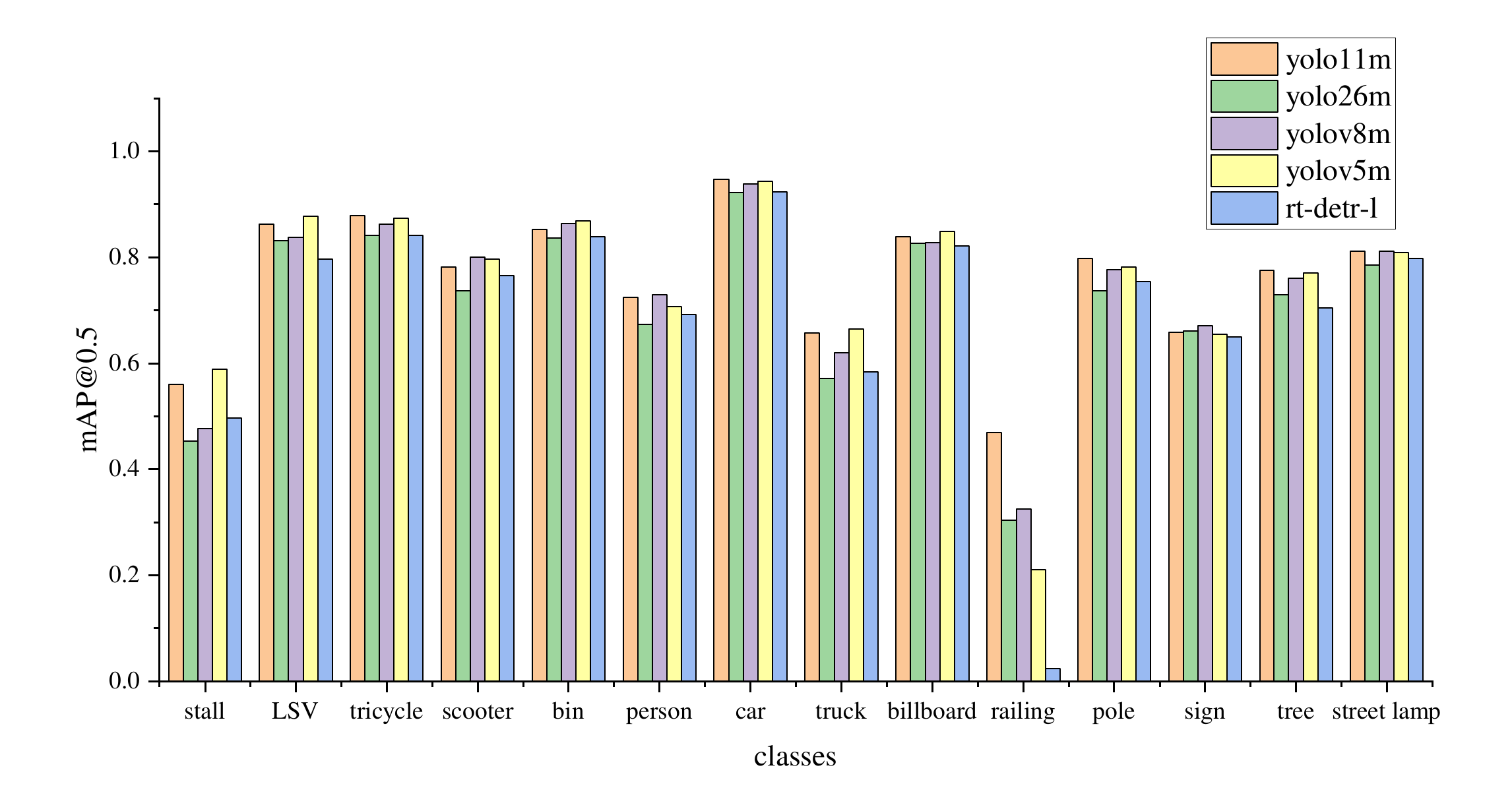}
    \caption{Per-class mAP@0.5 of different models under Configuration B. }
    \label{fig:placeholder}
\end{figure}

\begin{table*}[t]   
\centering
\caption{Per-class mAP@0.5 of different models under Configuration B.}
\resizebox{\textwidth}{!}{
    \begin{tabular}{l l l l l l l l l l l l l l l}
    \hline
     & stall & LSV & tricycle & scooter & bin & person & car & truck & billboard & railing & pole & sign & tree & street lamp \\
    \hline
    YOLO11m & 0.56 & 0.863 & 0.879 & 0.782 & 0.853 & 0.725 & 0.947 & 0.657 & 0.839 & 0.47 & 0.798 & 0.659 & 0.775 & 0.812 \\
    YOLO26m & 0.453 & 0.831 & 0.841 & 0.737 & 0.837 & 0.674 & 0.922 & 0.572 & 0.826 & 0.304 & 0.737 & 0.661 & 0.73 & 0.786 \\
    YOLOv8m & 0.477 & 0.838 & 0.863 & 0.8 & 0.864 & 0.73 & 0.938 & 0.62 & 0.828 & 0.325 & 0.777 & 0.671 & 0.761 & 0.812 \\
    YOLOv5mu & 0.589 & 0.877 & 0.874 & 0.797 & 0.869 & 0.707 & 0.944 & 0.665 & 0.849 & 0.211 & 0.782 & 0.655 & 0.771 & 0.809 \\
    RT-DETR-L & 0.497 & 0.797 & 0.841 & 0.765 & 0.839 & 0.692 & 0.923 & 0.584 & 0.822 & 0.024 & 0.754 & 0.65 & 0.705 & 0.798 \\
    \hline
    \end{tabular}%
}
\label{tab:perclass_map}
\end{table*}

\section{Conclusion }

In this paper, we systematically constructed a hybrid real-synthetic object detection dataset tailored for Chinese rural roads to address the data scarcity and generalization challenges faced by autonomous driving perception in complex Chinese rural traffic scenes. The performance of 13 mainstream object detection models was comprehensively evaluated on 517 real test images under three data configurations: all-real, 1:0.5 real-to-synthetic mixed, and 1:1 real-to-synthetic mixed. The experiments yielded the following conclusions.

First, the hybrid real-synthetic data strategy proves effective, and an optimal mixing ratio exists. A moderate amount of synthetic data can effectively improve object detection performance. Experimental results demonstrated that the 1:0.5 real-to-synthetic mixing configuration achieves the highest average mAP@0.5 for multiple models, with YOLO11m attaining the global best accuracy of 0.758 under this configuration. However, when the proportion of synthetic data becomes excessive, reaching a real-to-synthetic ratio of 1:1, the excessive domain bias offsets the gains brought by the increased data scale, leading to performance degradation on the real-scene test set.

Second, model architectures exhibit significant differences in their sensitivity to synthetic data. Medium-scale YOLO11 and YOLOv8 models can achieve considerable mAP@0.5 improvements with the 1:0.5 mixed data. In contrast, YOLOv5mu and the Transformer-based RT-DETR are more sensitive to the domain shift introduced by synthetic data. Notably, RT-DETR-L suffered a substantial performance drop after the incorporation of synthetic data. This indicated that the global attention mechanism is more susceptible to interference when confronted with distribution discrepancies between real and synthetic data. Consequently, when synthetic data are employed to assist training, strategy design should be tailored to the architectural characteristics of the model.

Moreover, model capacity is a key factor in coping with complex rural scenes. The experiments verified that, under unified hyperparameters, the detection accuracy of medium-scale models across all YOLO series comprehensively and consistently surpasses that of their small and nano counterparts. The numerous objects with highly variable shapes in Chinese rural traffic scenes necessitate high network parameter capacity for complex spatial feature modeling; overly small models are highly prone to representation bottlenecks.

Despite the remarkable progress achieved in detecting China-specific objects, the perception bottleneck for long-tail non-standard objects persists. While models exhibit excellent recognition rates on common objects and typical rural vehicles such as tricycles and LSVs—confirming the effectiveness of the proposed real-synthetic mixed dataset—existing mainstream detectors still face clear representation limitations for roadside stalls and elongated railings, regardless of whether they adopt CNN or Transformer architectures. To address these challenges, future work should focus on narrowing the domain gap between synthetic and real data through high-fidelity 3D assets, advanced lighting simulation, and generative AI techniques, while simultaneously optimizing detection architectures for long-tail non-standard features to enhance robustness under complex occlusions and background interference.

This study provides critical empirical guidance for model selection and synthetic data strategy design, facilitating the practical deployment of autonomous driving perception systems in rural areas.

\bibliography{IEEEabrv, conference_071817}

@inproceedings{geiger2012we,
  title={Are we ready for autonomous driving? the kitti vision benchmark suite},
  author={Geiger, Andreas and Lenz, Philip and Urtasun, Raquel},
  booktitle={2012 IEEE conference on computer vision and pattern recognition},
  pages={3354--3361},
  year={2012},
  organization={IEEE}
}

@inproceedings{redmon2016you,
  title={You only look once: Unified, real-time object detection},
  author={Redmon, Joseph and Divvala, Santosh and Girshick, Ross and Farhadi, Ali},
  booktitle={Proceedings of the IEEE conference on computer vision and pattern recognition},
  pages={779--788},
  year={2016}
}

@inproceedings{redmon2017yolo9000,
  title={YOLO9000: better, faster, stronger},
  author={Redmon, Joseph and Farhadi, Ali},
  booktitle={Proceedings of the IEEE conference on computer vision and pattern recognition},
  pages={7263--7271},
  year={2017}
}

@article{redmon2018yolov3,
  title={Yolov3: An incremental improvement},
  author={Redmon, Joseph and Farhadi, Ali},
  journal={arXiv preprint arXiv:1804.02767},
  year={2018}
}

@article{bochkovskiy2020yolov4,
  title={Yolov4: Optimal speed and accuracy of object detection},
  author={Bochkovskiy, Alexey and Wang, Chien-Yao and Liao, Hong-Yuan Mark},
  journal={arXiv preprint arXiv:2004.10934},
  year={2020}
}

@article{jocher2020yolov5,
  title={Yolov5 https://github. com/ultralytics/yolov5},
  author={Jocher, Glenn and Nishimura, K and Minerva, T and Vilari{\~n}o, R},
  journal={Accessed March},
  volume={7},
  pages={2021},
  year={2020}
}

@inproceedings{carion2020end,
  title={End-to-end object detection with transformers},
  author={Carion, Nicolas and Massa, Francisco and Synnaeve, Gabriel and Usunier, Nicolas and Kirillov, Alexander and Zagoruyko, Sergey},
  booktitle={European conference on computer vision},
  pages={213--229},
  year={2020},
  organization={Springer}
}

@inproceedings{zhao2024detrs,
  title={Detrs beat yolos on real-time object detection},
  author={Zhao, Yian and Lv, Wenyu and Xu, Shangliang and Wei, Jinman and Wang, Guanzhong and Dang, Qingqing and Liu, Yi and Chen, Jie},
  booktitle={Proceedings of the IEEE/CVF conference on computer vision and pattern recognition},
  pages={16965--16974},
  year={2024}
}

@article{hossain2025evaluating,
  title={Evaluating YOLO Architectures: Implications for Real-Time Vehicle Detection in Urban Environments of Bangladesh},
  author={Hossain, Ha Meem and Nath, Pritam and Mahi, Mahitun Nesa and Uddin, Imtiaz and Eiste, Ishrat Jahan and Ratul, Syed Nasibur Rahman and Mozumdar, Md Naim Uddin and Saad, Asif Mohammed and Hossain, MD},
  journal={arXiv preprint arXiv:2509.05652},
  year={2025}
}

@article{alimov2024domain,
  title={Domain Generalization in Autonomous Driving: Evaluating YOLOv8s, RT-DETR, and YOLO-NAS with the ROAD-Almaty Dataset},
  author={Alimov, Madiyar and Meiramkhanov, Temirlan},
  journal={arXiv preprint arXiv:2412.12349},
  year={2024}
}

@article{schoder2023first,
  title={First qualitative observations on deep learning vision model YOLO and DETR for automated driving in Austria},
  author={Schoder, Stefan},
  journal={arXiv preprint arXiv:2312.12314},
  year={2023}
}

@inproceedings{sun2017revisiting,
  title={Revisiting unreasonable effectiveness of data in deep learning era},
  author={Sun, Chen and Shrivastava, Abhinav and Singh, Saurabh and Gupta, Abhinav},
  booktitle={Proceedings of the IEEE international conference on computer vision},
  pages={843--852},
  year={2017}
}

@article{geiger2013vision,
  title={Vision meets robotics: The kitti dataset},
  author={Geiger, Andreas and Lenz, Philip and Stiller, Christoph and Urtasun, Raquel},
  journal={The international journal of robotics research},
  volume={32},
  number={11},
  pages={1231--1237},
  year={2013},
  publisher={Sage Publications Sage UK: London, England}
}

@inproceedings{yu2020bdd100k,
  title={Bdd100k: A diverse driving dataset for heterogeneous multitask learning},
  author={Yu, Fisher and Chen, Haofeng and Wang, Xin and Xian, Wenqi and Chen, Yingying and Liu, Fangchen and Madhavan, Vashisht and Darrell, Trevor},
  booktitle={Proceedings of the IEEE/CVF conference on computer vision and pattern recognition},
  pages={2636--2645},
  year={2020}
}

@inproceedings{caesar2020nuscenes,
  title={nuscenes: A multimodal dataset for autonomous driving},
  author={Caesar, Holger and Bankiti, Varun and Lang, Alex H and Vora, Sourabh and Liong, Venice Erin and Xu, Qiang and Krishnan, Anush and Pan, Yu and Baldan, Giancarlo and Beijbom, Oscar},
  booktitle={Proceedings of the IEEE/CVF conference on computer vision and pattern recognition},
  pages={11621--11631},
  year={2020}
}

@inproceedings{sun2020scalability,
  title={Scalability in perception for autonomous driving: Waymo open dataset},
  author={Sun, Pei and Kretzschmar, Henrik and Dotiwalla, Xerxes and Chouard, Aurelien and Patnaik, Vijaysai and Tsui, Paul and Guo, James and Zhou, Yin and Chai, Yuning and Caine, Benjamin and others},
  booktitle={Proceedings of the IEEE/CVF conference on computer vision and pattern recognition},
  pages={2446--2454},
  year={2020}
}

@article{cao2024semantic,
  title={Semantic segmentation network for unstructured rural roads based on improved SPPM and fused multiscale features},
  author={Cao, Xinyu and Tian, Yongqiang and Yao, Zhixin and Zhao, Yunjie and Zhang, Taihong},
  journal={Applied Sciences},
  volume={14},
  number={19},
  pages={8739},
  year={2024},
  publisher={MDPI}
}

@article{yao2025construction,
  title={Construction and enhancement of a rural road instance segmentation dataset based on an improved stylegan2-ada},
  author={Yao, Zhixin and Xi, Renna and Zhang, Taihong and Zhao, Yunjie and Tian, Yongqiang and Hou, Wenjing},
  journal={Sensors},
  volume={25},
  number={8},
  pages={2477},
  year={2025},
  publisher={MDPI}
}

@article{che2019d,
  title = {D\textsuperscript{2}-city: a large-scale dashcam video dataset of diverse traffic scenarios},
  author={Che, Zhengping and Li, Guangyu and Li, Tracy and Jiang, Bo and Shi, Xuefeng and Zhang, Xinsheng and Lu, Ying and Wu, Guobin and Liu, Yan and Ye, Jieping},
  journal={arXiv preprint arXiv:1904.01975},
  year={2019}
}

@article{wang2024m4sfwd,
  title={M4SFWD: A Multi-Faceted synthetic dataset for remote sensing forest wildfires detection},
  author={Wang, Guanbo and Li, Haiyan and Li, Peng and Lang, Xun and Feng, Yanling and Ding, Zhaisehng and Xie, Shidong},
  journal={Expert Systems with Applications},
  volume={248},
  pages={123489},
  year={2024},
  publisher={Elsevier}
}

@inproceedings{ye2022rope3d,
  title={Rope3d: The roadside perception dataset for autonomous driving and monocular 3d object detection task},
  author={Ye, Xiaoqing and Shu, Mao and Li, Hanyu and Shi, Yifeng and Li, Yingying and Wang, Guangjie and Tan, Xiao and Ding, Errui},
  booktitle={Proceedings of the IEEE/CVF Conference on Computer Vision and Pattern Recognition},
  pages={21341--21350},
  year={2022}
}

@article{zhu2021detection,
  title={Detection and tracking meet drones challenge},
  author={Zhu, Pengfei and Wen, Longyin and Du, Dawei and Bian, Xiao and Fan, Heng and Hu, Qinghua and Ling, Haibin},
  journal={IEEE transactions on pattern analysis and machine intelligence},
  volume={44},
  number={11},
  pages={7380--7399},
  year={2021},
  publisher={IEEE}
}

@article{voronin2024enhancing,
  title={Enhancing object detection accuracy in autonomous vehicles using synthetic data},
  author={Voronin, Sergei and Siddique, Abubakar and Iqbal, Muhammad},
  journal={arXiv preprint arXiv:2411.15602},
  year={2024}
}

@inproceedings{damian2023experimental,
  title={Experimental results on synthetic data generation in unreal engine 5 for real-world object detection},
  author={Damian, Alexandru and Filip, Claudiu and Nistor, Anamaria and Petrariu, Irina and Mariuc, C{\u{a}}t{\u{a}}lin and Stratan, Valentin},
  booktitle={2023 17th international conference on engineering of modern electric systems (emes)},
  pages={1--4},
  year={2023},
  organization={IEEE}
}

@article{kim2024experimental,
  title={Experimental study on using synthetic images as a portion of training dataset for object recognition in construction site},
  author={Kim, Jaemin and Wang, Ingook and Yu, Jungho},
  journal={Buildings},
  volume={14},
  number={5},
  pages={1454},
  year={2024},
  publisher={MDPI}
}

@inproceedings{de2025optimizing,
  title={Optimizing Object Detection for Maritime Search and Rescue: Progressive Fine-Tuning of YOLOv9 with Real and Synthetic Data.},
  author={de Lima, Luciano Netto and de Alcantara Andrade, Fabio Augusto and Djenouri, Youcef and Pfeiffer, Carlos and Moura, Marcos},
  booktitle={ICAART (3)},
  pages={209--216},
  year={2025}
}

@article{samak2025sim2real,
  title={Sim2Real Diffusion: Leveraging Foundation Vision Language Models for Adaptive Automated Driving},
  author={Samak, Chinmay and Samak, Tanmay and Li, Bing and Krovi, Venkat},
  journal={IEEE Robotics and Automation Letters},
  volume={11},
  number={1},
  pages={177--184},
  year={2025},
  publisher={IEEE}
}

@inproceedings{marcus2025synth,
  title={Synth It Like KITTI: Synthetic Data Generation for Object Detection in Driving Scenarios},
  author={Marcus, Richard and Vogel, Christian and Jatzkowski, Inga and Knoop, Niklas and Stamminger, Marc},
  booktitle={International Conference on Robotics, Computer Vision and Intelligent Systems},
  pages={414--432},
  year={2025},
  organization={Springer}
}

@article{delussu2024synthetic,
  title={Synthetic data for video surveillance applications of computer vision: A review},
  author={Delussu, Rita and Putzu, Lorenzo and Fumera, Giorgio},
  journal={International Journal of Computer Vision},
  volume={132},
  number={10},
  pages={4473--4509},
  year={2024},
  publisher={Springer}
}

@article{ljungqvist2023object,
  title={Object Detector Differences when Using Synthetic and Real Training Data: MG Ljungqvist et al.},
  author={Ljungqvist, Martin Georg and Nordander, Otto and Skans, Markus and Mildner, Arvid and Liu, Tony and Nugues, Pierre},
  journal={SN computer science},
  volume={4},
  number={3},
  pages={302},
  year={2023},
  publisher={Springer}
}

@article{remmas2025pcgod,
  title={PCGOD: Enhancing Object Detection With Synthetic Data for Scarce and Sensitive Computer Vision Tasks},
  author={Remmas, Walid and Lints, Martin and Uudm{\"a}e, Jaak Joonas},
  journal={IEEE Access},
  year={2025},
  publisher={IEEE}
}

@inproceedings{venkatesh2025volucapture,
  title={VoluCapture: Multi-View Synthetic Data Capture in Unreal Engine},
  author={Venkatesh, Sandeep Bangalore and Su, Guan-Ming},
  booktitle={2025 IEEE 8th International Conference on Multimedia Information Processing and Retrieval (MIPR)},
  pages={551--554},
  year={2025},
  organization={IEEE}
}

@inproceedings{d2025syndra,
  title={Syndra: Synthetic dataset for railway applications},
  author={D’Amico, Gianluca and Nesti, Federico and Rossolini, Giulio and Marinoni, Mauro and Sabina, Salvatore and Buttazzo, Giorgio},
  booktitle={2025 IEEE/CVF Winter Conference on Applications of Computer Vision (WACV)},
  pages={3437--3446},
  year={2025},
  organization={IEEE}
}

\end{document}